\pdfoutput=1
\PassOptionsToPackage{table}{xcolor}

\documentclass[10pt, logo, twocolumn, copyright, nonumbering]{nvidiatechreport}

\usepackage{natbib}
\usepackage[capitalize,nameinlink]{cleveref}
\usepackage{latexsym}
\usepackage{multirow}
\usepackage{array}
\usepackage{csvsimple}
\usepackage{makecell}
\usepackage{placeins}
\usepackage{inconsolata}
\usepackage{xurl}

\graphicspath{{figures/}}
\newcolumntype{L}{>{\raggedright\arraybackslash}X}

\definecolor{lightgreen}{HTML}{f0f7e8}
\definecolor{rowshade}{HTML}{ffffff}

\newcommand{\uncval}[2]{\makecell[r]{#1\\[-0.25ex]\scriptsize$\pm$#2}}
\newcommand{\uncvalbf}[2]{\makecell[r]{\textbf{#1}\\[-0.25ex]\scriptsize$\pm$#2}}

\newcommand{\orangechange}[1]{#1}
\newcommand{\revtable}[1]{#1}

\newif\ifshowacks
\showackstrue

\title{\textsc{SHERLOC}: Structured Diagnostic Localization for Code Repair Agents}

\author[1,2]{Hovhannes Tamoyan}
\author[1]{Sean Narenthiran}
\author[1]{Erik Arakelyan}
\author[2,3]{Mira Mezini}
\author[1]{Boris~Ginsburg}

\affil[1]{NVIDIA, Santa Clara, CA 95051, USA}
\affil[2]{TU Darmstadt, Darmstadt, Germany}
\affil[3]{hessian.AI \& National Research Center for Applied Cybersecurity ATHENE}


\begin{document}
\begin{abstract}
LLM agents solve repository-level coding tasks through multi-turn tool use, but utilize half their budget on locating faults before editing.
Dedicated localization frameworks have emerged, yet are still evaluated as file retrieval rather than actionable diagnosis, producing locations without the diagnostic context a repair agent needs.
We introduce \textsc{SHERLOC} (\emph{Structured Hypothesis-driven Exploration and Reasoning for Localization}), a training-free framework pairing a reasoning LLM with compact repository tools and self-recovery, without fine-tuning or multi-agent orchestration.
\textsc{SHERLOC} reaches state-of-the-art localization across model scales: $84.33\%$ $\texttt{accuracy@}1$ on \textsc{SWE-Bench Lite} and $81.27\%$ $\texttt{recall@}1$ on \textsc{SWE-Bench Verified}; at $\sim30$B parameters, it matches or outperforms other agentic methods.
Injecting our locations and diagnostic findings into repair agents yields an average $+5.95$~pp resolve-rate gain from the best \textsc{SHERLOC} result per setting on \textsc{SWE-Bench Verified}. \textsc{SHERLOC} cuts localization and total tokens by $36.7\%$ and $23.1\%$ on average.\footnotemark
\end{abstract}
\maketitle
\footnotetext{\url{https://github.com/NVIDIA-NeMo/Skills/tree/main/recipes/sherloc}}


\section{Introduction}
\label{sec:intro}

Repository-level code repair increasingly relies on LLM agents that interleave extended reasoning with structured tool calls over a codebase~\citep{wei2022chain, yao2022react, schick2023toolformer}.
A canonical benchmark for evaluating these agents is \textsc{SWE-Bench}~\citep{jimenez2024swebench}, built from GitHub issues and their resolving pull requests: from a bug report and a repository snapshot, the agent must propose and verify a bug-fixing code patch.
Yet how agents allocate their interaction budget within this repair loop remains underexplored.
In a systematic study spanning $5$ LLMs and $2$ agent frameworks, we find that agents spend on average $18.5$ turns ($48\%$ of total interaction) locating faults before their first patch, consuming over $320$k tokens per instance (\Cref{app:localization_efficiency}).
This makes localization both a performance bottleneck and a dominant compute cost, consuming context and interaction budget that could otherwise support patch construction.

\begin{figure}[t]
    \centering
    \includegraphics[width=\columnwidth]{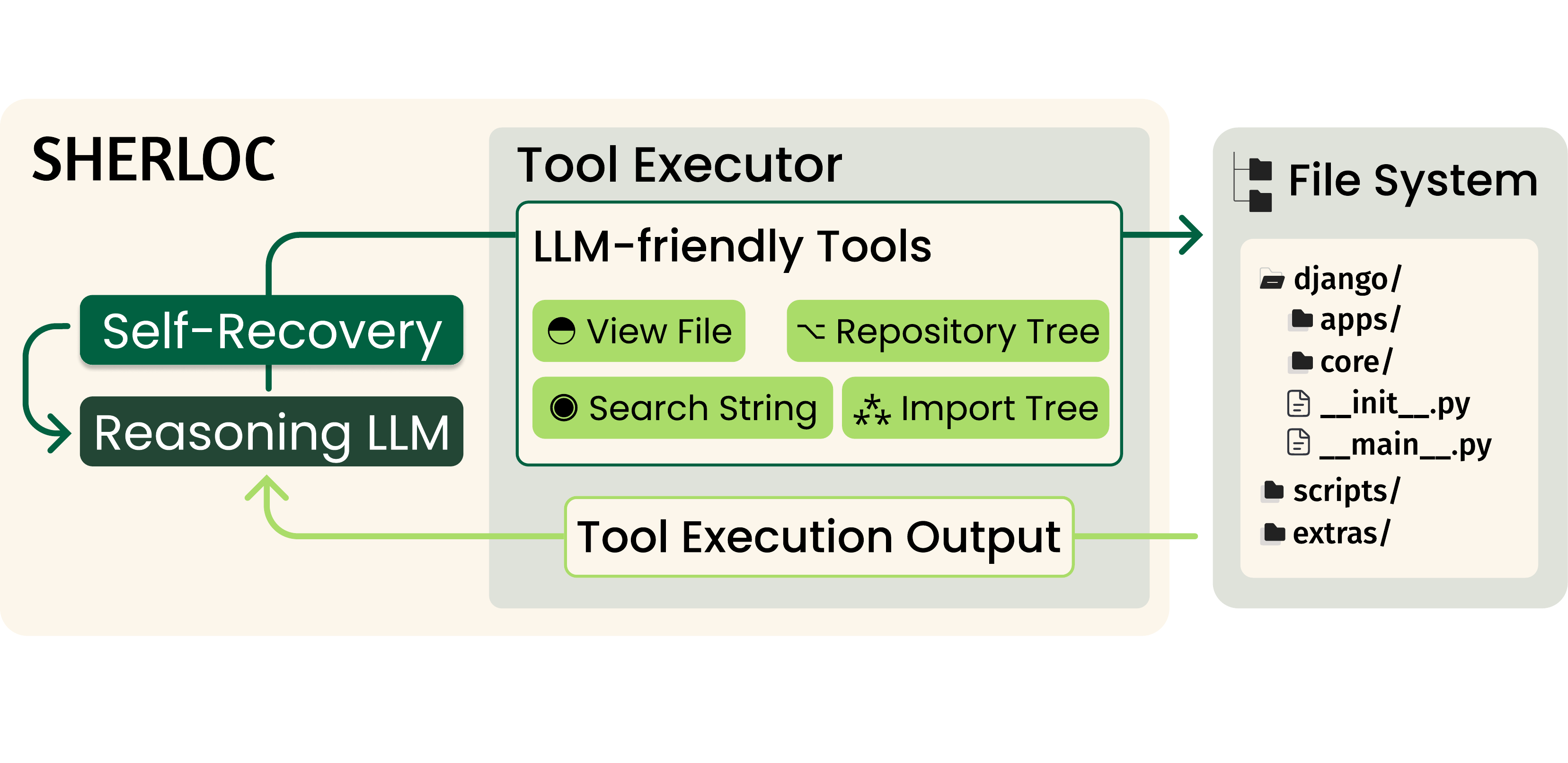}
    \caption{\textbf{Overview of \textsc{SHERLOC}}.
    A reasoning LLM interacts with a tool executor over four LLM-friendly tools, with a self-recovery layer correcting failures.
    \textsc{SHERLOC} achieves state-of-the-art file-level localization on \textsc{SWE-Bench Lite} and \textsc{Verified}. In downstream code repair, its best result per setting averages a $5.95$~pp resolve-rate gain; On average \textsc{SHERLOC} cuts localization and total tokens by $36.7\%$ and $23.1\%$.}
    \label{fig:teaser}
\end{figure}

\begin{figure*}[t!]
  \centering
  \begin{minipage}[t]{0.49\textwidth}
      \centering
      \vspace{0pt}
      \includegraphics[width=\linewidth]{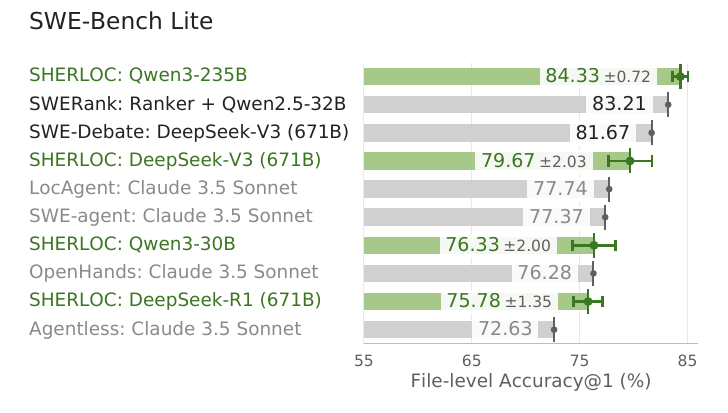}
  \end{minipage}\hfill
  \begin{minipage}[t]{0.49\textwidth}
      \centering
      \vspace{0pt}
      \includegraphics[width=\linewidth]{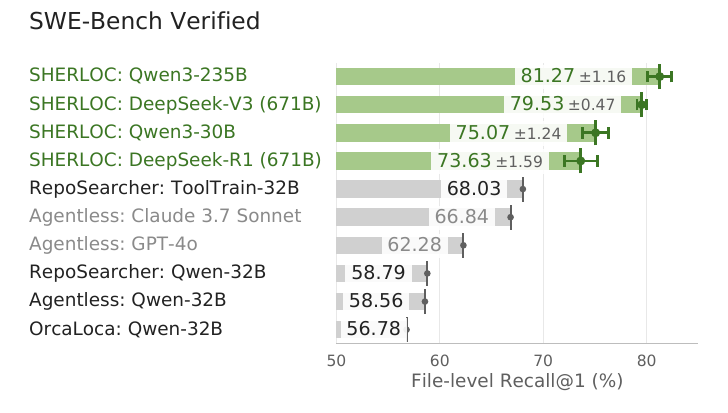}
  \end{minipage}
  \caption{\textbf{Localization performance across benchmarks}.
  Left: file-level accuracy@$1$ on \textsc{SWE-Bench Lite}; right: file-level recall@$1$ on \textsc{SWE-Bench Verified}.
  Labels use compact model names for readability; proprietary model labels appear in light gray.
  \textsc{SHERLOC} results are means over $3$ seeds ($\pm$ std); prior systems report single runs.
  Backbone sizes differ across systems; \Cref{ssec:cross_model} provides a controlled cross-model analysis.
  }
  \label{fig:combined_results}
\end{figure*}

Most localization methods retrieve faulty files and functions~\citep{zhou2012buglocator, reddy2025sweranksoftwareissuelocalization, yu2025orcaloca}; some, e.g., OrcaLoca and SWE-Debate~\mbox{\citep{li2025swe}}, also produce free-form report text.
This view is necessary but incomplete: a file path tells the agent where to look, but not why, and a file path without a root-cause hypothesis leaves downstream repair under-specified.
We hypothesize that explicitly producing candidate locations, root-cause analysis, and actionable solution guidance yields a richer, more transferable localization, and ask: \textbf{(RQ1)} Can a single reasoning LLM with a compact, structured tool interface match task-specifically trained localizers without fine-tuning or multi-agent orchestration? \textbf{(RQ2)} When do its locations and findings transfer to code-repair agents, and how does the resolve rate and token cost shift?

We introduce \textsc{SHERLOC} (\emph{Structured Hypothesis-driven Exploration and Reasoning for Localization}), a training-free framework coupling a reasoning LLM with a compact suite of LLM-friendly repository tools (file viewing, code search, repository-tree inspection, and import-graph navigation; \Cref{fig:teaser}) and lightweight self-recovery mechanisms (context truncation, loop detection, malformed-tool-call repair, final-turn synthesis).
For each predicted location, \textsc{SHERLOC} emits a structured diagnostic \emph{finding}.

For \textbf{(RQ1)}, we evaluate \textsc{SHERLOC} across model families on \textsc{SWE-Bench Lite} and \textsc{SWE-Bench Verified}.
It reaches state-of-the-art file-level localization on both benchmarks and remains competitive at a matched $\sim30$B scale, without supervised fine-tuning, reinforcement learning, or multi-agent debate.
Ablations show that the tool suite, self-recovery mechanisms, and reasoning mode all contribute, while implicit-knowledge controls show that the gains persist when file paths in the issue are masked.

Strong localization does not automatically translate to better repository-level issue resolution.
For \textbf{(RQ2)}, we inject our locations and findings into two code-repair frameworks, OpenHands~\mbox{\citep{wang2025openhandsopenplatformai}} and SWE-Agent~\mbox{\citep{yang2024sweagentagentcomputerinterfacesenable}}, across $5$ repair-agent backbones.
Across the $10$ model-framework pairs, the per-pair best of unfiltered \textsc{SHERLOC} and quality-filtered \textsc{SHERLOC} (QF \textsc{SHERLOC}) averages a $+5.95$~pp resolve-rate gain; deployable, unfiltered \textsc{SHERLOC} averages $+4.39$~pp, with $36.7\%$ and $23.1\%$ fewer localization and total tokens.

\begin{figure}[tbp]
  \centering
  \includegraphics[width=0.85\columnwidth]{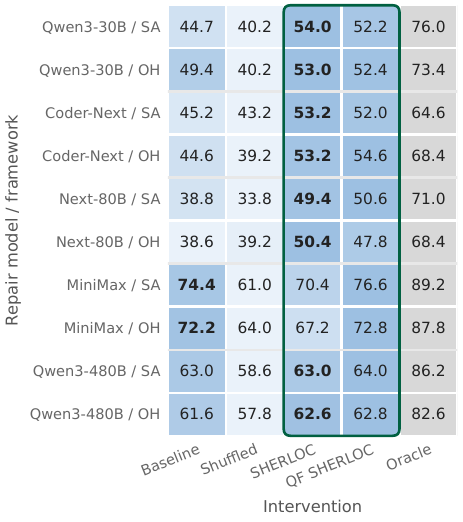}
  \caption{\textbf{Downstream resolve rates of LLM repair agents on \textsc{SWE-Bench Verified}}.
  Each cell reports resolve rate in percentages; within each row, non-Oracle conditions are shaded blue from light (worst) to dark (best).
  In row labels, SA denotes SWE-Agent and OH denotes OpenHands.
  Oracle GPT-$5.2$ (light gray) is an upper reference, not a deployable intervention.
  Bold values indicate the best deployable result in each row, comparing Baseline with \textsc{SHERLOC}.
  Paired differences and $95\%$ confidence intervals appear in \Cref{tab:e2e_paired_uncertainty}.}
  \label{fig:e2e_quality}
\end{figure}

\begin{figure}[tbp]
  \centering
  \includegraphics[width=\columnwidth]{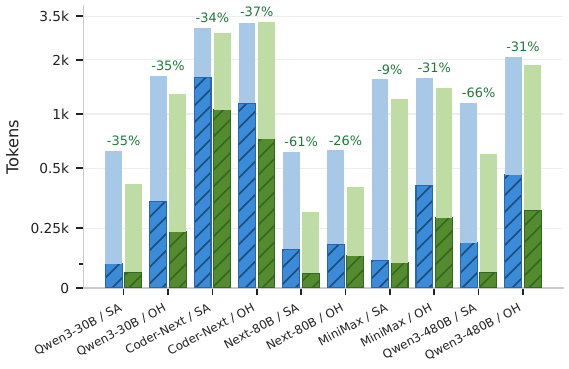}
  \caption{\textbf{Search-efficiency gains from \textsc{SHERLOC} findings}.
  Localization-phase (hatched) and full-run (solid) per-instance token costs: baseline (blue) vs.\ \textsc{SHERLOC} (green); the delta above each pair is the localization-token change relative to baseline.
  In x-axis labels, SA denotes SWE-Agent and OH denotes OpenHands.
  The largest reductions in localization-token usage occur with bigger models, even when their resolve rates are already saturated.
  Full numbers in \Cref{app:localization_efficiency}.}
  \label{fig:loc_efficiency}
\end{figure}

\textsc{SHERLOC} improves or preserves the resolve-rate point estimate in eight of ten settings and reduces localization tokens in all ten (\Cref{fig:e2e_quality,fig:loc_efficiency}).
QF \textsc{SHERLOC} mitigates the only observed case of negative transfer.
Thus, transfer is positive on average but quality-mediated.

Our main contributions are:
\begin{itemize}
    \item \textbf{A training-free, tool-augmented localization framework.} \textsc{SHERLOC} pairs a single reasoning LLM with a compact suite of LLM-friendly repository tools and lightweight self-recovery, without task-specific training or multi-agent orchestration, and reaches state-of-the-art file-level localization on \textsc{SWE-Bench Lite} ($84.33\%$ accuracy@$1$) and \textsc{SWE-Bench Verified} ($81.27\%$ recall@$1$).
    \item \textbf{Diagnostic \emph{findings} as structured localization context.} Alongside candidate locations, \textsc{SHERLOC} produces structured findings with $5$ fields (location explanation, root cause, solution idea, dependencies, testing impact) for each predicted location.
    \item \textbf{A structure-agnostic localization metric.} We propose chunk-level metrics over (start, end) line-number spans (coverage recall, precision, and average tightness) as a structure-agnostic alternative to function-, class-, or module-level evaluation, since not every patch target sits within a syntactic unit.
    \item \textbf{Cross-family transfer with validity controls.} \textsc{SHERLOC} transfers across model families ($79.53\%$ recall@$1$ with DeepSeek-V3-0324); implicit-knowledge controls show its gains persist when issue paths are masked ($79.96\%$ vs.\ $81.27\%$), isolating active exploration from parametric familiarity with public \textsc{SWE-Bench} repositories.
    \item \textbf{Dual-axis transfer to downstream code repair.} Injecting \textsc{SHERLOC}'s findings into existing code-repair agents yields an average $+5.95$~pp resolve-rate gain from the best \textsc{SHERLOC} result per setting; \textsc{SHERLOC} cuts localization and total tokens by $36.7\%$ and $23.1\%$ across five repair backbones and two frameworks; a retrospective QF \textsc{SHERLOC} analysis shows that excluding unreliable findings can mitigate negative transfer.
\end{itemize}

\section{Related Work}

\paragraph{Tool-augmented LLM reasoning.}
\textsc{SHERLOC} builds on reasoning-augmented tool use~\citep{wei2022chain,yao2022react,schick2023toolformer} and on test-time scaling that adapts reasoning compute to problem difficulty~\citep{wu2025thoughtcalibration,zhao2025t2}.
Unlike prior localization systems built on supervised fine-tuning, reinforcement learning, or multi-agent orchestration, ours uses a single LLM with a fixed, structured tool set.

\paragraph{LLM-based code localization.}
Recent LLM-based methods explore repositories interactively or reason over code graphs: SWE-Debate adjudicates fault-propagation traces through multi-agent debate~\citep{li2025swe}; OrcaLoca uses priority-based action scheduling and distance-aware context pruning~\citep{yu2025orcaloca}; CoSIL and LocAgent guide LLM search through code-graph structure~\citep{jiang2025cosilissuelocalizationllmdriven,chen2025locagent}; RepoSearcher/ToolTrain uses tool-integrated reinforcement learning~\citep{ma2025toolintegratedreinforcementlearningrepo}.
\textsc{SHERLOC} differs in design: no training, no auxiliary structures, a compact set of LLM-friendly repository actions, and self-recovery mechanisms that stabilize extended-thinking tool use.
Whereas prior work typically targets file- or function-level entity locations, sometimes with prescriptive modification plans for downstream patching~\citep{li2025swe}, our method additionally produces structured \emph{diagnostic} findings alongside each location.

\paragraph{Issue-based retrieval and ranking.}
Information-retrieval localization treats the bug report as a query and source files as documents.
BugLocator~\citep{zhou2012buglocator} ranks files using textual similarity and prior bug reports; BLUiR~\citep{saha2013bluir} incorporates structured source-code information; AmaLgam~\citep{wang2014amalgam} combines version history, similar reports, and structure.
Recent neural rankers such as SWERank learn issue--code similarity with task-specific training~\citep{reddy2025sweranksoftwareissuelocalization}.
These methods output ranked locations rather than diagnostic reports; \textsc{SHERLOC} instead frames localization as active repository reasoning yielding structured findings.

\paragraph{Fault localization with execution signals.}
Classical fault localization assumes access to executable tests or dynamic traces: \textbf{spectrum-based} methods rank by suspiciousness from coverage and pass/fail statistics~\citep{jones2002visualization,abreu2007accuracy}; \textbf{mutation-based} methods inject synthetic faults and observe test-outcome changes~\citep{moon2014ask,papadakis2015metallaxis}; learning-based extensions train neural models over coverage or graph features~\citep{li2019deepfl,lou2021boosting}; AutoFL uses LLM tool calls but still requires at least one failing test~\citep{kang2024autofl}.
Unlike these execution-dependent methods, \textsc{SHERLOC} operates from the issue description and repository snapshot alone, without executing tests or observing coverage.

\paragraph{Localization granularity.}
Existing localization studies report at module, file, class, or function level~\citep{kang2024autofl,chen2025locagent}.
For repository-level repair, this granularity is not always aligned with the target edit: patches may modify class-level declarations, imports, configuration logic, or coordinated regions across functions.
Several recent function-level evaluations sidestep this mismatch by discarding $26$ of the $300$ \textsc{SWE-Bench Lite} instances that lack an existing function or class target~\citep{suresh2025cornstack,chen2025locagent,reddy2025sweranksoftwareissuelocalization}.
We therefore complement file-level metrics with structure-agnostic chunk-level metrics over line ranges (coverage recall, precision, and average tightness) that represent predictions and ground truth as (start, end) line-number spans, making no assumption about which syntactic unit contains the patch target while retaining all benchmark instances.

\paragraph{Issue-resolution agents and benchmark validity.}
Issue-resolution systems target patch correctness directly: Agentless~\citep{xia2024agentlessdemystifyingllmbasedsoftware} and AutoCodeRover~\citep{zhang2024autocoderoverautonomousprogramimprovement} decompose the task into localization, patch generation, and validation; agentic systems such as SWE-Agent and OpenHands integrate navigation, editing, and execution in a single loop~\citep{yang2024sweagentagentcomputerinterfacesenable,wang2025openhandsopenplatformai}.
We instead evaluate localization directly and, in separate experiments, inject our findings into existing repair agents to test downstream transfer.
Because high file-identification accuracy on \textsc{SWE-Bench} can partly reflect pretraining familiarity~\citep{liang2025swebenchillusion}, we apply implicit-knowledge, shuffled, and masked controls to separate benchmark-specific signal from tool-assisted diagnostic reasoning.


\section{Method}
\label{sec:method}

We formulate localization as an iterative, tool-mediated reasoning problem: given a natural-language issue report and a repository snapshot, the system must identify code regions likely to require modification and produce a diagnostic finding, not merely a ranked location, that explains the suspected root cause and solution direction.
\textsc{SHERLOC} implements this with four components (\Cref{fig:teaser}): a reasoning LLM, a deterministic executor mediating repository access, a compact suite of LLM-friendly tools, and lightweight self-recovery.
The model alternates between reasoning and structured actions until it has enough evidence to emit final findings and locations.
Its hypothesis state is the accumulated conversation history, including prior observations; no external memory store is maintained.

\begin{figure}[tbp]
    \centering
    \includegraphics[width=\columnwidth]{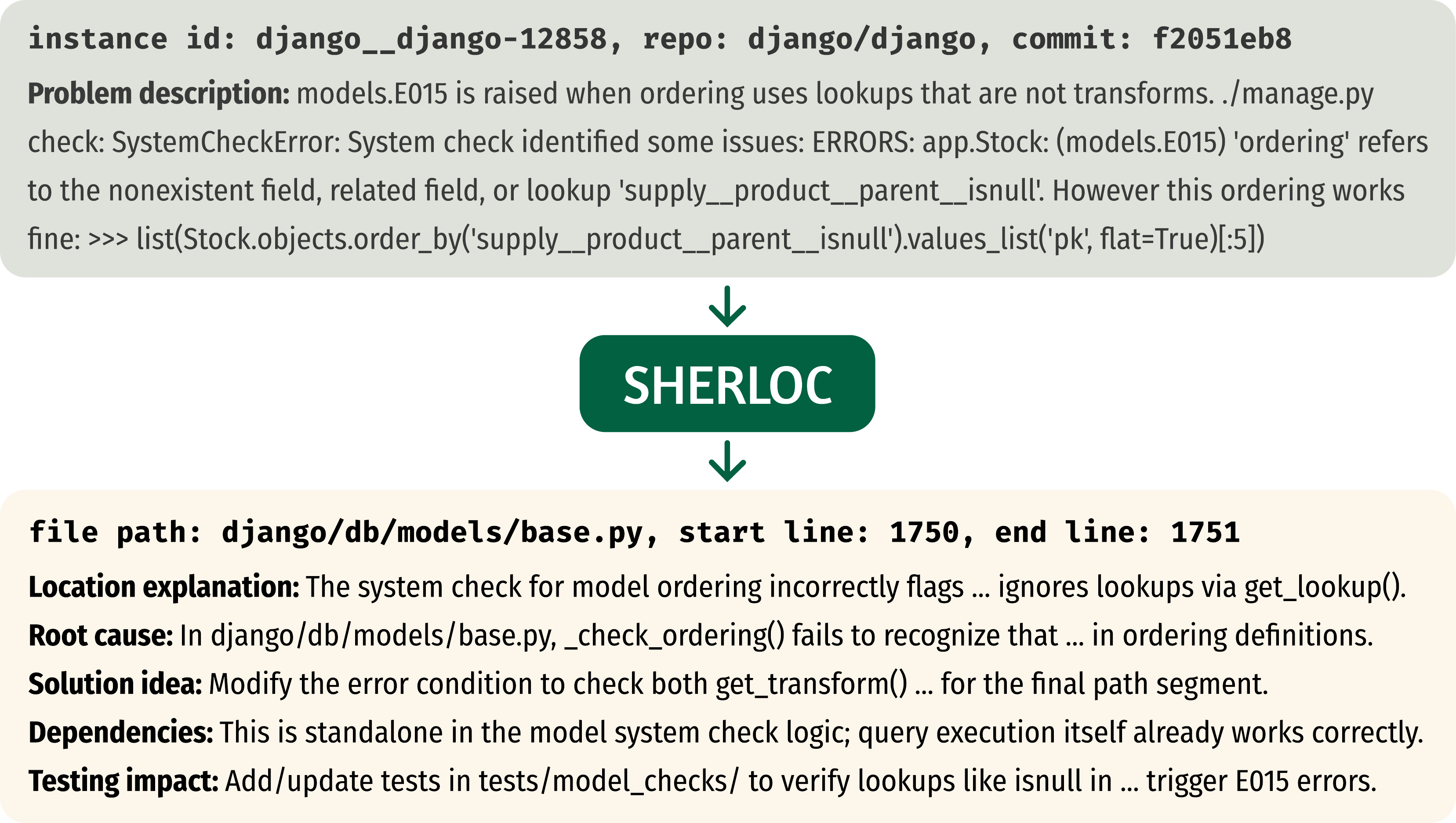}
    \caption{\textbf{Example \textsc{SHERLOC} input and output}.
    Given a \textsc{SWE-Bench Verified} Django issue, \textsc{SHERLOC} predicts a line span in \texttt{django/db/models/base.py} and emits a structured diagnostic finding with the location explanation, root cause, solution idea, dependencies, and testing impact.}
    \label{fig:example_in_out}
\end{figure}

\paragraph{Design principle: reasoning-native tool use.}
Rather than synthesizing arbitrary shell commands or Python scripts for repository actions, the model selects from a small fixed set of tools, each with a structured schema for execution (e.g., file path and optional line range) and bounded, model-readable observations such as snippets, paths, line ranges, and dependency summaries.
This keeps repository exploration legible while preserving a deterministic execution boundary: the model chooses the action, but the executor validates and runs it, avoiding incidental command construction or shell-debug failure modes that prior work has shown to derail agent trajectories under raw shell access~\citep{yang2024sweagentagentcomputerinterfacesenable}.
It also lets us use reasoning models, which excel at multi-step deliberation but are not necessarily trained for external tool APIs, without task-specific tool-use fine-tuning.

\subsection{Initialization and Context}
Before prompting, we construct a filtered repository view that removes directories unlikely to contain production logic, such as documentation, build artifacts, dependency folders, and version-control metadata (e.g., \texttt{/docs}, \texttt{/.git}).
The initial prompt combines this filtered tree with the issue description, tool descriptions, and required output format, giving the model a global map of the project without loading source files.

\subsection{Tool Suite}
The suite is limited by design: enough to expose relevant repository evidence, but not so broad as to invite open-ended software-engineering actions.
\textbf{View File} inspects a file, optionally restricted to a line range, with fuzzy path matching that suggests corrections for near-miss names.\quad{}\textbf{Codebase Search} performs repository-wide literal search and returns matching snippets with surrounding context.\quad{}\textbf{Repository Tree} displays the filtered file hierarchy (like shell \textsc{tree}), letting the model regain global context.\quad{}\textbf{Connected Tree} summarizes import relationships (direct and reverse) to follow dependencies across modules.

\subsection{Iterative Interaction Loop}
At the core of \textsc{SHERLOC} is a bounded interaction loop of at most $20$ turns; at each turn the model either makes a tool call or terminates with final locations and findings.
The loop has four steps.
\textbf{Reasoning and action selection}: the LLM receives the issue, repository context, and previous observations, then decides whether additional evidence is needed and, if so, emits one structured action.\quad{}\textbf{Action parsing}: a parser extracts a tool call or a final answer; final answers contain a \texttt{findings} block ($5$ fields: location explanation, root cause, solution idea, dependencies, testing impact; see \Cref{fig:example_in_out,app:example_findings}) and a \texttt{locations} block.\quad{}\textbf{Tool execution}: the executor runs the action and appends the observation to the conversation history.\quad{}\textbf{Termination}: the loop ends when a valid final answer is parsed or when the step budget is exhausted, in which case a final-turn prompt forces synthesis from gathered evidence.

\subsection{Self-Recovery Mechanisms}
We add recovery for common failure modes in multi-turn tool use (\Cref{app:self_recovery_mechanisms}): context management (truncating older observations), loop detection (warnings for repeated unproductive calls), implicit tool-call recovery (parsing malformed but unambiguous requests), response-length management (re-prompting when generation approaches safe limits), and final-turn prompting (forcing synthesis when the step budget is exhausted).

\begin{figure}[tbp]
    \centering
    \includegraphics[width=\columnwidth]{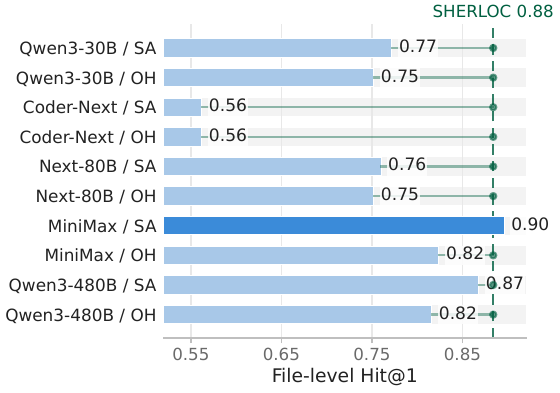}
    \caption{\textbf{Localization headroom by downstream LLM agent}.
    File-level Hit@$1$ of each code repair LLM agent's own localization (blue) vs.\ \textsc{SHERLOC} (green dashed line, $0.88$).
    The gap between each bar and the \textsc{SHERLOC} line is the agent's headroom; weaker agents have larger headroom and gain more from external findings.
    Full numbers in \Cref{app:localization_efficiency}.}
    \label{fig:loc_headroom}
\end{figure}


\section{Results}
\label{sec:results}

We evaluate \textsc{SHERLOC} along four axes: (i) the quality of its predicted code locations and (ii) the robustness of those predictions to benchmark contamination and to LLM backbone choice (addressing \textbf{RQ1}; \Cref{ssec:main_results,ssec:implicit_knowledge,ssec:cross_model}); and (iii) the transfer of its outputs to downstream code repair, gated by (iv) the quality of its diagnostic findings under an LLM judge (addressing \textbf{RQ2}; \Cref{ssec:e2e}).
We report file-level metrics (precision, recall, F1, exact match, hit@$1$) over the predicted file set and structure-agnostic chunk-level metrics (coverage recall, precision, average tightness) over (start, end) line-number spans; definitions in \Cref{app:full-metrics}.

\subsection{Localization Performance and Efficiency}
\label{ssec:main_results}
We evaluate our framework across \textsc{SWE-Bench Lite}~\mbox{\citep{jimenez2024swebench}} and \textsc{SWE-Bench Verified}~\mbox{\citep{chowdhury2024swebenchverified}} using $4$ backbone models and $3$ seeds.
\Cref{fig:combined_results} compares against leading localizers including SWE-Debate~\citep{li2025swe}, SWERank~\citep{reddy2025sweranksoftwareissuelocalization}, LocAgent~\citep{chen2025locagent}, SWE-agent~\citep{yang2024sweagentagentcomputerinterfacesenable}, OpenHands~\citep{wang2025openhandsopenplatformai}, and Agentless~\citep{xia2024agentlessdemystifyingllmbasedsoftware}.
Following prior localization work~\citep{li2025swe,reddy2025sweranksoftwareissuelocalization,chen2025locagent}, we report file-level accuracy@$1$ as the fraction of instances whose top-ranked predicted file is a gold file and recall@$1$ as the mean fraction of gold files recovered by the top-ranked prediction.

On \textsc{SWE-Bench Lite}, our framework with Qwen3-235B-A22B-Thinking-2507 (Qwen3-235B)~\citep{yang2025qwen3technicalreport,qwen2025qwen3235bthinking2507} achieves \textbf{$84.33\pm0.72\%$} accuracy@$1$ in $5.0$ turns on average; the strongest plotted prior result is the fine-tuned SWERank retrieve-and-rerank pipeline ($83.21\%$).
At the chunk level, it reaches $39.14\%$ coverage recall and $44.54\%$ precision.

On \textsc{SWE-Bench Verified}, the best configuration (Qwen3-235B) achieves \textbf{$81.27\pm1.16\%$} recall@$1$ with $4.7$ average turns, a $13.2$~pp improvement over the previous state of the art, RepoSearcher / ToolTrain-$32$B~\citep[$68.03\%$;][]{ma2025toolintegratedreinforcementlearningrepo}.
At a matched $\sim 30$B scale and without task-specific training, Qwen3-30B-A3B-Thinking-2507 reaches $75.07\pm1.24\%$ recall@$1$, outperforming ToolTrain-$32$B by $7.0$~pp.
At the chunk level, it reaches $41.28\%$ coverage recall and $53.47\%$ precision.
Because prior systems do not report chunk-level metrics, these span scores primarily characterize how tightly \textsc{SHERLOC}'s predicted regions cover the gold edit spans rather than serving as a direct baseline comparison.

DeepSeek-V3-0324~\citep{deepseekai2024deepseekv3technicalreport,deepseekai2025deepseekv30324} reaches $79.53\pm0.47\%$ on Verified, confirming cross-family transfer.
Full per-backbone numbers (including chunk-level metrics and tool-engagement) appear in \Cref{tab:sherloc_full_scores}; the complete metric grid is in \Cref{app:full-metrics}.

At the trajectory level, the search remains compact while scaling with problem difficulty: turns rise from $4.22$ (easy) to $5.05$/$5.21$ (medium/hard) and total tokens from $21.3$k to $32.8$k/$35.1$k (\Cref{sec:token_usage_analysis,app:difficulty_appendix}).

\begin{figure}[t]
  \centering
  \includegraphics[width=\columnwidth]{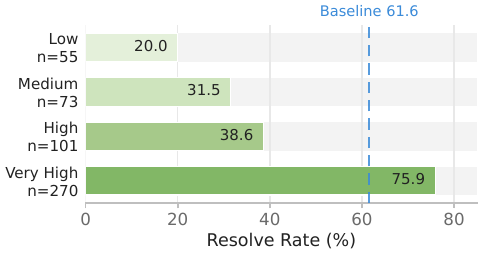}
  \caption{\textbf{Resolve rate by composite finding-quality bucket}.
  \textsc{SWE-Bench Verified} / \textsc{OpenHands} / Qwen3-Coder-480B-A35B-Instruct.
  Buckets: Low~[$1.0$,$2.0$], Medium~($2.0$,$3.0$], High~($3.0$,$4.0$], Very~High~($4.0$,$5.0$].
  Very-high-quality findings resolve at $75.9\%$; low-quality findings actively underperform the $61.6\%$ baseline.}
  \label{fig:quality_resolve}
\end{figure}

\subsection{Component Ablations}
\label{ssec:ablation_overview}
To keep the full component grid tractable, we ablate each component of our framework and show the impact of each separate tool and mechanism on a fixed sample of $100$ instances drawn uniformly from the SWE-Gym development set~\citep{pan2025trainingsoftwareengineeringagents} (full breakdown in \Cref{app:ablation}).
Removing any single tool or self-recovery mechanism degrades localization, with \texttt{View File} ($-7.0$~pp F1) and the final-turn prompt ($-5.0$~pp F1) as the largest contributors, indicating that both code inspection and synthesis pressure are central.

\subsection{Implicit Knowledge vs.\ Tool-Assisted Reasoning}
\label{ssec:implicit_knowledge}

Because \textsc{SWE-Bench} repositories are public and widely represented in pretraining corpora, localization accuracy can reflect both active reasoning and implicit knowledge of familiar codebases~\citep{liang2025swebenchillusion}.
We treat this as a validity concern and conduct a controlled masking study with Qwen3-235B on \textsc{SWE-Bench Verified}, progressively removing repository access and identifying information (\Cref{tab:implicit_controls}).
Even when tools, the repository tree, and all file and module paths in the issue are masked, the model still achieves \textbf{$57.86\%$ recall@$1$} from the issue text alone, often inferring the correct file from error messages, API names, and domain conventions.
We interpret this primarily as repository familiarity with widely used foundational libraries, not as direct evidence that individual benchmark instances are memorized.
Manually inspecting a random sample of $50$ successful instances from this setting confirms $2$ patterns: reliance on (i) well-known APIs and libraries, and (ii) distinctive error messages that uniquely identify a module.
Quantitatively (\Cref{tab:per_repo_contamination}), implicit recall concentrates in popular projects ($85.3\%$ on scikit-learn, $87.5\%$ on requests), while less common ones like pylint and seaborn fall to $33.3\%$.

Crucially, this baseline familiarity does not explain \textsc{SHERLOC}'s full performance.
When tools and the repository tree are retained but explicit paths are removed from the issue text, \textsc{SHERLOC} reaches \textbf{$79.96\%$} recall@$1$, only $1.3$~pp below the unmasked full system (\textbf{$81.27\%$}).
The $\approx22$~pp gap between this masked-issue setting and the heavily-masked text-only setting estimates the contribution of active repository exploration beyond parametric familiarity.

\subsection{Backbone Generalization and Tool Engagement}
\label{ssec:cross_model}

To separate the \textsc{SHERLOC} method from the backbone, we evaluate $5$ models from $2$ families: Qwen3-235B-A22B-Thinking-2507, Qwen3-30B-A3B-Thinking-2507, and Qwen3-30B-A3B-Instruct-2507~\citep{yang2025qwen3technicalreport,qwen2025qwen3235bthinking2507,qwen2025qwen330binstruct2507,qwen2025qwen330bthinking2507}; DeepSeek-V3-0324~\citep{deepseekai2024deepseekv3technicalreport,deepseekai2025deepseekv30324}; and DeepSeek-R1-0528~\citep{deepseekai2025deepseekr1,deepseekai2025deepseekr10528}.
Runs use identical code, prompts, tools, and self-recovery; only the backbone changes.
Each uses $3$ seeds on \textsc{SWE-Bench Verified}; recall, mean turns, and zero-tool rate are in \Cref{tab:sherloc_full_scores}.

First, \textsc{SHERLOC} \textbf{transfers across model families}: DeepSeek-V3-0324 reaches $79.53\%$ recall, only $1.7$~pp below the Qwen3-235B ceiling and $11.5$~pp above the best published baseline~\citep[ToolTrain-$32$B, $68.03\%$;][]{ma2025toolintegratedreinforcementlearningrepo}.
Second, \textbf{tool engagement matters}: DeepSeek-R1-0528 makes $41\%$ fewer tool calls and $32\%$ fewer turns than Qwen3-235B, with $10\%$ producing locations without any tool call; those zero-tool instances reach $69.3\%$ recall vs.\ $73.6\%$ on its tool-using subset, so active exploration helps even strong reasoning models.
Third, \textbf{scale alone is not enough}: Qwen3-30B outperforms the larger R1-0528 ($75.07\%$ vs.\ $73.63\%$) while taking the most turns ($7.2$ on average, vs.\ $4.7$ for Qwen3-235B and $3.2$ for R1-0528).
R1-0528's under-exploration is consistent with DeepSeek's lack of a marked system-role token: the prompt is delivered as bare preamble, reducing tool-use compliance.
The Qwen3-30B-Thinking vs.\ Qwen3-30B-Instruct ablation (\Cref{app:thinking_vs_instruct}) reinforces this: reasoning is critical for sustaining the multi-turn protocol.

\subsection{LLM-as-Judge Scoring of Findings}
\label{ssec:finding_quality}
How reliable are the diagnostic findings that \textsc{SHERLOC} produces? \textsc{SHERLOC} produced locations and findings for $499$ of the $500$ \textsc{SWE-Bench Verified} instances, with one instance yielding no prediction. We score these $499$ findings with an LLM-as-judge.
We use GPT-$5.2$~\citep{openai2025gpt5} as a ground-truth-patch-conditioned judge that rates each finding from $1$ to $5$ on three dimensions: \textbf{root-cause correctness}, \textbf{location accuracy}, and \textbf{solution actionability} (prompt details in \Cref{app:prompts}; full quality-scoring breakdown in \Cref{app:finding_quality}).
The composite quality score is the mean of these three dimensions, and we call a finding \emph{high-quality} when this composite is $\geq 4.0$.
The scores are right-skewed: $270$ ($54\%$) fall in Very High ($> 4.0$), $101$ ($20\%$) in High ($3.0$ to $4.0$), $73$ ($15\%$) in Medium ($2.0$ to $3.0$), and $55$ ($11\%$) in Low ($\leq 2.0$) (\Cref{tab:quality_distribution}).
The $\geq 4.0$ threshold therefore retains $317$/$499$ scored findings ($63.5\%$).


\subsection{Downstream Transfer to Code Repair Task}
\label{ssec:e2e}

Does better localization translate into better downstream issue resolution? We measure code-repair \emph{resolve rate} (fraction of generated patches that pass the held-out tests) on all $500$ \textsc{SWE-Bench Verified} instances~\mbox{\citep{chowdhury2024swebenchverified}}, using $2$ agentic frameworks, SWE-Agent~\mbox{\citep{yang2024sweagentagentcomputerinterfacesenable}} and OpenHands~\mbox{\citep{wang2025openhandsopenplatformai}}, across $6$ setups (\Cref{fig:e2e_quality}; full grid in \Cref{tab:e2e_quality_full}): \textbf{Baseline} (vanilla agent execution); \textbf{Masked} (baseline with file paths and repository names masked, extending the implicit-knowledge control from \Cref{ssec:implicit_knowledge}); \textbf{Shuffled findings} (a \textsc{SHERLOC} location and finding from a different random instance); \textbf{\textsc{SHERLOC} findings} (\textsc{SHERLOC}'s predicted locations and findings); \textbf{QF \textsc{SHERLOC}} (retrospectively selects the \textsc{SHERLOC} outcome for scored findings with judge score $\geq 4.0$ and the Baseline outcome below the threshold); and \textbf{Oracle GPT-$5.2$} (upper-bound finding conditioned on the ground-truth patch).

\textbf{Transfer gains depend on model capability.}
\textsc{SHERLOC} findings improve or preserve the resolve-rate point estimate in eight of ten settings and reduce localization tokens in all ten (\Cref{fig:e2e_quality,fig:loc_efficiency}).
\emph{Smaller and mid-sized models gain most from \textsc{SHERLOC} findings.}
For Qwen3-Coder-30B-A3B~\citep{qwen2025qwen3coder30b}, SWE-Agent rises from $44.7\%$ to $54.0\%$ and OpenHands from $49.4\%$ to $53.0\%$; Qwen3-Coder-Next~\citep{cao2026qwen3codernext} and Qwen3-Next-80B-A3B-Instruct~\citep{qwen2025qwen3next} show $+8$ to $+12$~pp improvements across both frameworks (\Cref{fig:e2e_quality}).
These models have low baseline resolve and large localization headroom (Qwen3-Coder-Next reaches only $0.56$ Hit@$1$ vs.\ our $0.88$; \Cref{fig:loc_headroom}), so even moderate-quality findings close a real gap.
\emph{QF \textsc{SHERLOC} mitigates the only observed case of negative transfer.}
For MiniMax-M2.5~\citep{minimax2026m25}, \textsc{SHERLOC} lowers resolve rate by $4$ to $5$~pp across the two frameworks (\Cref{fig:e2e_quality}).
QF \textsc{SHERLOC} raises these results to $76.6\%$ with SWE-Agent and $72.8\%$ with OpenHands, but neither differs significantly from Baseline (\Cref{app:e2e_paired_uncertainty}).
\emph{Token efficiency improves uniformly.}
Across all $5$ repair models and $2$ frameworks, \textsc{SHERLOC} findings reduce localization tokens by $9$ to $66\%$ ($36.7\%$ on average) and total tokens by $23.1\%$ on average (\Cref{fig:loc_efficiency,app:localization_efficiency}), giving the agent a targeted starting point and freeing context and interaction budget for repair.
The largest cell-level efficiency gain is Qwen3-Coder-480B-A35B~\citep{qwen2025qwen3coder} / SWE-Agent, where injecting findings cuts localization tokens from $188.9$k to $64.6$k ($-66\%$) while leaving resolve rate unchanged at $63.0\%$, showing that even for already-saturated repair agents the standalone localizer pays for itself in search cost alone.
The masked and shuffled controls remain close to baseline (\Cref{tab:baseline_vs_sherloc_localization}), so gains come from instance-relevant diagnostic guidance, not surface leakage or unrelated findings.

\subsection{\texorpdfstring{Understanding Downstream Transfer}{Understanding Downstream Transfer}}
\label{ssec:e2e_analysis}

\begin{figure}[h!]
    \centering
    \includegraphics[width=\columnwidth]{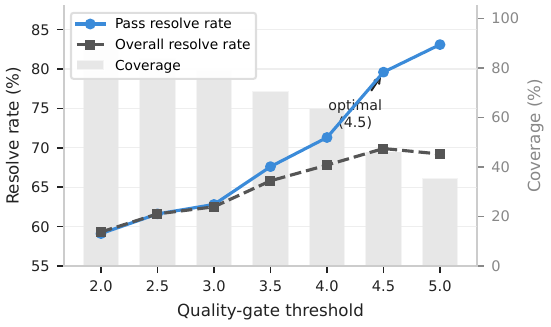}
    \caption{\textbf{Quality-threshold sweep}.
    Lines (left axis): Pass resolved (blue) is resolve rate among instances meeting the threshold; Overall resolved (dashed) is the filtered estimate with $61.6\%$ baseline fallback for instances with rejected findings.
    Bars (right axis): coverage among the $499$ scored findings.
    Stricter thresholds raise resolve rate while shrinking coverage; Overall resolved peaks at $4.5$.}
    \label{fig:threshold_sensitivity}
\end{figure}

\textbf{Finding quality is associated with downstream success.}
On the Qwen3-Coder-480B-A35B / OpenHands cell, very-high-quality findings resolve at $75.9\%$, compared with $20.0\%$ for low-quality ones, and composite score correlates with repair outcome at $r{=}0.45$ (\Cref{fig:quality_resolve}).
Using the $\geq 4.0$ threshold defined in \Cref{ssec:finding_quality}, we divide instances by the quality of their \textsc{SHERLOC} findings.
Above the threshold, conditioning the repair agent on these findings raises resolve rate from $74.8\%$ under Baseline to $80.4\%$; below the threshold, it lowers resolve rate from $38.5\%$ to $31.3\%$.
QF \textsc{SHERLOC} therefore retains the \textsc{SHERLOC} outcome only above the threshold and falls back to Baseline otherwise, mitigating negative transfer in \Cref{ssec:e2e}.
The pre-specified $4.0$ threshold is the highest tested half-point cutoff retaining a majority of scored findings ($317/499$; $63.5\%$) and is within $2.1$~pp of the best retrospective estimate at $4.5$ (\Cref{tab:threshold_sensitivity,fig:threshold_sensitivity}).

\textbf{Patch-free verification provides a deployable confidence signal.}
Because QF \textsc{SHERLOC} relies on a ground-truth-patch-conditioned judge, it is retrospective.
As an optional deployable safeguard for repair agents whose validation results indicate negative transfer, our patch-free verifier evaluates the canonical finding using the issue description and five auxiliary traces.
Across $499$ findings, its confidence score correlates with retrospective quality at $r{=}0.553$; at threshold $4.0$, it selects $281$ findings with $79.0\%$ precision and $70.0\%$ recall.
This experiment measures agreement with retrospective quality labels, not downstream repair under verifier-based selection.

\textbf{Solution content carries that signal.}
Two independent lenses converge: by per-dimension correlation on the Qwen3-Coder-480B-A35B / OpenHands cell, solution actionability has the strongest individual association with repair success ($r{=}0.46$, vs.\ $r{=}0.41$ for root-cause and $r{=}0.36$ for location accuracy); by causal field ablation on Qwen3-Coder-30B-A3B / SWE-Agent (\Cref{tab:decomposition}), removing the \emph{solution idea} field produces the largest single-field drop while all other reduced variants still beat the baseline.
The transferable signal therefore comes from correct root-cause and solution content, not from merely seeing a \textsc{SHERLOC}-formatted hint.

\textbf{Chunk coverage predicts downstream repair success.}
To separate chunk coverage from file identification, we restrict the analysis to the $371$ instances for which \textsc{SHERLOC} identifies all and only the correct files, so every file metric is $100\%$.
Among these, $144$ instances have complete coverage of the gold edit chunks and $227$ have incomplete coverage.
Across the ten \textsc{SHERLOC} repair settings, the complete coverage group resolves at $82.5\%$, compared with $59.9\%$ for the incomplete coverage group, a $+22.6$~pp difference.
The gap is positive in every setting and ranges from $+18.8$ to $+27.4$~pp (\Cref{tab:chunk_coverage_repair}).
Thus, chunk coverage provides information about downstream repair success even when file identification is perfect.
Chunk metrics complement rather than replace standard file metrics.

\textbf{Findings help most on multi-file bugs.}
On the $71$ Verified instances requiring $2$ or more file changes, \textsc{SHERLOC} findings improve resolve rate by $+5.6$~pp ($32.4\%$ vs.\ $26.8\%$).

Finally, we separate accuracy from efficiency transfer by injecting findings from the weaker Qwen3-30B \textsc{SHERLOC} (Hit@$1$ $0.786$ vs.\ $0.884$; \Cref{tab:qwen30_findings_localization}).
These weaker findings no longer improve Qwen3-Coder-30B-A3B but still improve Qwen3-Coder-Next (SWE-Agent $45.2\%$$\to$$49.2\%$; OpenHands $44.6\%$$\to$$45.4\%$; \Cref{tab:e2e_quality_qwen30_findings}), and still reduce localization tokens by $16$ to $32\%$ ($24.5\%$ on average) and total tokens by $1$ to $17\%$ ($7.5\%$ on average) across all $4$ cells (\Cref{tab:localization_efficiency_qwen30_findings}).
Localizer quality thus governs accuracy transfer; even imperfect findings reduce search cost by anchoring initial exploration.

\section{Conclusion}
\label{sec:conclusion}

Our work on \textsc{SHERLOC} shows that structured diagnostic output, not just location retrieval, is the operative unit of useful localization.
For \textbf{RQ1}, a single reasoning LLM with compact repository tools and self-recovery can reach state-of-the-art file-level localization without task-specific fine-tuning or multi-agent orchestration: $84.33\%$ accuracy@$1$ on \textsc{SWE-Bench Lite} and $81.27\%$ recall@$1$ on \textsc{SWE-Bench Verified}, with matched-scale performance competitive with other agentic methods.
For \textbf{RQ2}, injecting these locations and diagnostic findings into existing code-repair agents yields an average $+5.95$~pp resolve-rate gain from the best \textsc{SHERLOC} result per setting; On average \textsc{SHERLOC} cuts localization and total tokens by $36.7\%$ and $23.1\%$ across $5$ repair backbones and $2$ frameworks. A retrospective QF \textsc{SHERLOC} analysis shows that excluding unreliable findings can mitigate negative transfer.

We conclude that localization should be evaluated not just by file or function retrieval but by diagnostic actionability: \textbf{a correct file with a misleading diagnosis still misleads the downstream agent.}
Beyond bug localization, structured findings may benefit repository-scale tasks such as test generation, regression triage, refactoring, and migration planning, where intermediate reasoning matters as much as the final action.


\section*{Limitations}

\paragraph{Implicit-knowledge confound.}
Our implicit-knowledge controls show that $\approx58\%$ of localization recall on \textsc{SWE-Bench Verified} is achievable from issue text alone, with strong per-repository concentration: scikit-learn and requests reach $85$--$88\%$ masked recall while pylint and seaborn fall to $33\%$ (\Cref{app:per_repo}).
Headline \textsc{SWE-Bench} localization numbers, ours included, therefore partly reflect LLM pretraining familiarity with widely distributed open-source code rather than transferable code reasoning.
Our masked-issue control ($79.96\%$ with tools retained) bounds but does not eliminate this confound; fully isolating pretraining familiarity would require evaluation on demonstrably unseen repositories.

\paragraph{Benchmark and framework scope.}
Code-repair transfer is measured on \textsc{SWE-Bench Verified} through two repair frameworks (\textsc{SWE-Agent}, \textsc{OpenHands}) and five repair backbones.
Generalization to private or demonstrably unseen repositories, other benchmarks such as \textsc{SWE-Bench Multimodal}~\mbox{\citep{yang2025swebenchmultimodal}} and multilingual variants, newer benchmark snapshots, additional agent frameworks, and non-Python repositories (JavaScript, Java, C++) remains open.
The \texttt{Connected Tree} tool relies on a thin language-specific import parser that must be adapted for each supported language; the rest of \textsc{SHERLOC} is language-agnostic.

\paragraph{Compute cost of the headline configuration.}
Our best numbers use Qwen3-235B-A22B-Thinking-2507 with up to $20$ reasoning turns per instance, so the headline localization rows are not directly comparable on serving cost to the fine-tuned $32$B baselines we cite.
The matched-scale $30$B row ($75.07\%$ on \textsc{Verified}, $+7.0$~pp over ToolTrain-$32$B) is the most cost-comparable point of evidence for the method itself; the $+6.2$~pp from $235$B over $30$B partly reflects model scale.

\paragraph{Cross-model prompt sensitivity.}
Identical prompts elicit different tool-use behavior across model families: DeepSeek-R1-0528, which does not support marked system roles, exhibits $10\%$ zero-tool shortcutting under the same prompt text (\Cref{ssec:cross_model}).
Systematic evaluation of commercial models such as GPT, Claude, and Gemini remains future work; they can be connected through a model/API adapter without changing the localization workflow, with the required integration points documented in the released code.
Picking a backbone with system-prompt support sidesteps this; otherwise, reaching the reported numbers on a new backbone may require model-specific prompt adaptation rather than a strictly drop-in deployment.

\paragraph{Retrospective QF \textsc{SHERLOC} and patch-free verification.}
The QF \textsc{SHERLOC} analysis uses GPT-$5.2$~\citep{openai2025gpt5} as an external judge that is shown the ground-truth patch.
The judge-independent cells in \Cref{fig:e2e_quality} (baseline, masked, shuffled, \textsc{SHERLOC}, Oracle GPT-$5.2$) provide non-judge evidence for the $2$ transfer regimes, but the per-instance reliability analysis and the threshold sweep in \Cref{app:threshold_sensitivity} depend on this external supervision.
The QF \textsc{SHERLOC} row should therefore be read as an analysis-time selection experiment.
Our patch-free verifier removes the ground-truth requirement, but we evaluate only its agreement with retrospective quality labels, not downstream repair under verifier-based selection.

\section*{Reproducibility Statement}

\textsc{SHERLOC} is implemented using the NeMo-Skills framework~\citep{nvidia2024nemoskills}.
The exact prompts (system prompt, tool descriptions, final-turn prompt, judge prompt) and inference parameters are provided in \Cref{app:prompts}.
All open-weight models used in this work are publicly available. Localization uses Qwen3-235B-A22B-Thinking-2507~\mbox{\citep{qwen2025qwen3235bthinking2507}}, Qwen3-30B-A3B-Thinking-2507~\mbox{\citep{qwen2025qwen330bthinking2507}}, Qwen3-30B-A3B-Instruct-2507~\mbox{\citep{qwen2025qwen330binstruct2507}}, DeepSeek-V3-0324~\mbox{\citep{deepseekai2025deepseekv30324}}, and DeepSeek-R1-0528~\mbox{\citep{deepseekai2025deepseekr10528}}. Downstream repair uses Qwen3-Coder-30B-A3B~\mbox{\citep{qwen2025qwen3coder30b}}, Qwen3-Coder-Next~\mbox{\citep{cao2026qwen3codernext}}, Qwen3-Next-80B-A3B-Instruct~\mbox{\citep{qwen2025qwen3next}}, MiniMax-M2.5~\mbox{\citep{minimax2026m25}}, and Qwen3-Coder-480B-A35B~\mbox{\citep{qwen2025qwen3coder480b}}. The LLM-as-judge quality scorer, GPT-$5.2$~\mbox{\citep{openai2025gpt5}}, is proprietary and accessed through its public API. Model names and versions are listed above, and the sampling parameters for every run are given in the \emph{Sampling parameters} paragraph below.
Evaluation is conducted on the public \textsc{SWE-Bench Lite} and \textsc{SWE-Bench Verified} benchmarks using their official evaluation harnesses.

\paragraph{Computing infrastructure and budget.}
All experiments are inference-only and run on internal compute clusters of NVIDIA H$100$ ($80$\,GB) and A$100$ ($80$\,GB) GPUs.
Backbones are served via SGLang~\citep{zheng2024sglang} for localization and vLLM~\citep{kwon2023vllm} for downstream code repair.
Approximately $7{,}900$ GPU-hours were used for the downstream repair experiments ($5$ repair backbones $\times$ $2$ frameworks $\times$ $6$ conditions), approximately $2{,}500$ GPU-hours for the multi-backbone, multi-seed localization evaluations, and a few hundred GPU-hours for the component and reasoning-mode ablations and implicit-knowledge controls.
The Qwen3-235B-A22B-Thinking-2507 localization runs used $16$ GPUs for serving and averaged approximately $107$ GPU-hours per $500$ tasks, or $0.21$ GPU-hours per task.
Standard \textsc{SHERLOC} requires one localization run per issue; the optional patch-free verifier additionally requires five auxiliary localization runs and one verifier call.

\paragraph{Sampling parameters.}
Every \textsc{SHERLOC} localization run uses temperature $= 0.99$, top-$p = 0.95$, top-$k = 0$ (disabled), a per-call generation cap of $81{,}920$ tokens, and at most $20$ interaction turns.
All downstream code-repair backbones use temperature $= 0.7$, top-$p = 0.8$, and top-$k = 20$ under the stock \textsc{SWE-Agent} and \textsc{OpenHands} configurations.
The LLM-as-judge quality scorer (GPT-$5.2$) uses temperature $= 0.1$ and a $300$-token output cap.
The prompts and remaining implementation details are in \Cref{app:prompts}.

\ifshowacks
\section*{Acknowledgments}

We thank Somshubra Majumdar, Vahid Noroozi, Wasi Uddin Ahmad, Nikolai Ludwig, Mehrzad Samadi, Aleksander Ficek, and Siddhartha Jain (NVIDIA) for valuable discussions and feedback that helped shape this work.
The work benefited from funding by the DFG (German Research Foundation) under the Excellence Strategy, EXC-3057.
\fi

\bibliography{references}

\clearpage
\appendix
\crefname{section}{Appendix}{Appendices}
\Crefname{section}{Appendix}{Appendices}

\section*{Part I: Localization Analyses}

\section{Component Ablations: Tools and Self-Recovery}
\label{app:ablation}

\Cref{tab:tool_ablation,tab:self_recovery_ablation} evaluate the localization agent on the same $100$ uniformly sampled SWE-Gym development issues. We ablate one component at a time while holding the base model, prompts, sampling setup, and all other tools or self-recovery mechanisms fixed.

The sample covers $10$ of $11$ repositories, with repository shares differing from the full development split by at most $4.3$~pp. SWE-Gym follows the SWE-Bench instance-extraction and execution-based validation pipelines on separate Python repositories~\citep{pan2025trainingsoftwareengineeringagents}, providing a comparable development setting. We therefore interpret the ablation as evidence for qualitative component trends; exact effect sizes may vary on the full distribution.

File-level metrics (precision, recall, F1, exact match, and set accuracy) are computed over the predicted file set; chunk-level metrics (coverage recall and average tightness) are computed over (start, end) line-number spans. Parenthesized values are absolute-point deltas relative to the corresponding full-system row in the same table.

Across both ablations, the largest performance drops come from removing \texttt{View File} (tool suite) and the final-turn prompt (self-recovery), confirming that code inspection and forced final-turn synthesis are the two most load-bearing components.

\subsection{Reasoning-Mode Ablation: Thinking vs. Instruct}
\label{app:thinking_vs_instruct}

To isolate the contribution of extended reasoning, we compare \textbf{Qwen3-30B-A3B-Thinking-2507} against its non-thinking counterpart \textbf{Qwen3-30B-A3B-Instruct-2507}, which shares the same parameter count, architecture, prompts, and tools.
The Instruct model achieves only $10.2\%$ recall on Verified and $11.7\%$ on Lite, versus $74.0\%$ and $76.7\%$ for the Thinking model: a \textbf{$\sim65$~pp gap}.
The failure is not gradual degradation but near-total collapse: $87\%$ of Instruct samples fail to produce any valid output (success rate $13\%$), with the model averaging $0.2$ predicted locations per instance versus $1.3$ for the Thinking variant.
Without extended chain-of-thought, the model cannot sustain the multi-turn tool-use protocol that \textsc{SHERLOC} requires.
This ablation, controlled within a single model family, shows that thinking-mode reasoning is critical for this multi-turn exploration and diagnosis protocol.

\begin{table}[!htbp]
\centering
\scriptsize
\setlength{\tabcolsep}{2pt}
\caption{\textbf{Tool-suite ablation on $100$ SWE-Gym development issues}.
Each row removes one localization tool while holding the base model, prompt, self-recovery settings, and evaluation set fixed.
File-level columns report precision, recall, F1, exact match, and set accuracy; chunk-level columns report coverage recall and tightness.
Parenthesized values are absolute-point deltas from the all-tools full system.}
\label{tab:tool_ablation}
\resizebox{\columnwidth}{!}{%
\begin{tabular}{l ccc cc cc}
\toprule
\multirow{2}{*}{\textbf{Ablated Tool}} &
\multicolumn{3}{c}{\textbf{File-Level}} &
\multicolumn{2}{c}{\textbf{File-Level (set)}} &
\multicolumn{2}{c}{\textbf{Chunk-Level}} \\
\cmidrule(lr){2-4} \cmidrule(lr){5-6} \cmidrule(lr){7-8}
& \textbf{Prec.} & \textbf{Rec.} & \textbf{F1} & \textbf{Exact} & \textbf{Acc.} & \textbf{Cov.Rec.} & \textbf{Tight.} \\
\midrule
\rowcolor{rowshade}
\textbf{All Tools} & \textbf{77.1} & \textbf{58.23} & \textbf{63.19} & \textbf{37.0} & \textbf{56.31} & \textbf{27.49} & \textbf{27.97} \\
\midrule
w/o View File & \textbf{69.66} (-7.4) & \textbf{51.70} (-6.5) & \textbf{56.19} (-7.0) & 35.0 (-2.0) & \textbf{49.95} (-6.36) & \textbf{22.32} & \textbf{19.40} \\
\rowcolor{rowshade}
w/o Codebase Search & 71.73 (-5.3) & 53.42 (-4.8) & 57.70 (-5.4) & \textbf{32.0} (-5.0) & 50.59 (-5.72) & 21.75 & 20.42 \\
w/o Repo Tree & 76.33 (-0.7) & 56.88 (-1.3) & 62.34 (-0.8) & 38.0 (+1.0) & 55.96 (-0.35) & 28.20 & 26.02 \\
\rowcolor{rowshade}
w/o Connected Tree & 71.50 (-5.6) & 54.65 (-3.5) & 58.63 (-4.5) & 35.0 (-2.0) & 52.05 (-4.26) & 26.50 & 23.41 \\
\bottomrule
\end{tabular}
}
\end{table}

\begin{table}[!htbp]
\centering
\scriptsize
\setlength{\tabcolsep}{2pt}
\caption{\textbf{Self-recovery ablation on $100$ SWE-Gym development issues}.
Each row removes one recovery/control mechanism while holding the base model, tool suite, prompt, and evaluation set fixed.
Columns match \Cref{tab:tool_ablation}; parenthesized values are absolute-point deltas from the full system.}
\label{tab:self_recovery_ablation}
\resizebox{\columnwidth}{!}{%
\begin{tabular}{l ccc cc ccc}
\toprule
\multirow{2}{*}{\textbf{Ablated Component}} &
\multicolumn{3}{c}{\textbf{File-Level}} &
\multicolumn{2}{c}{\textbf{File-Level (set)}} &
\multicolumn{3}{c}{\textbf{Chunk-Level}} \\
\cmidrule(lr){2-4} \cmidrule(lr){5-6} \cmidrule(lr){7-9}
& \textbf{Prec.} & \textbf{Rec.} & \textbf{F1} & \textbf{Exact} & \textbf{Acc.} & \textbf{Cov.Rec.} & \textbf{Tight.} & \textbf{Prec.} \\
\midrule
\rowcolor{rowshade}
\textbf{All (full system)} & \textbf{77.1} & \textbf{58.23} & \textbf{63.19} & \textbf{37.0} & \textbf{56.31} & \textbf{27.49} & \textbf{27.97} & \textbf{49.22} \\
\midrule
w/o Implicit Tool Detection & 74.0 (-3.1) & 53.78 (-4.4) & 59.14 (-4.0) & 36.0 (-0.1) & 52.63 (-3.6) & 25.94 & \textbf{21.78} & 49.37 \\
\rowcolor{rowshade}
w/o Loop Detection & 76.66 (-0.4) & 56.74 (-1.4) & 61.78 (-1.4) & 36.0 (-0.1) & 54.67 (-1.6) & 26.52 & 30.86 & 52.38 \\
w/o Context Management & 73.33 (-3.7) & 55.16 (-3.0) & 59.84 (-3.3) & 39.0 (+2.0) & 54.05 (-2.2) & 26.18 & 28.04 & \textbf{48.02} \\
\rowcolor{rowshade}
w/o Response Length Mgmt. & 74.26 (-2.8) & 56.53 (-1.7) & 60.62 (-2.5) & 35.0 (-2.0) & 53.71 (-2.6) & 26.92 & 27.95 & 52.58 \\
w/o Final Turn Prompt & \textbf{72.5} (-4.6) & \textbf{54.08} (-4.1) & \textbf{58.19} (-5.0) & \textbf{31.0} (-6.0) & \textbf{50.75} (-5.5) & 27.76 & 23.66 & 49.30 \\
\bottomrule
\end{tabular}
}
\end{table}

\FloatBarrier

\section{Per-Backbone Localization Metrics}
\label{app:full-metrics}

\Cref{tab:sherloc_full_scores} reports the complete evaluation of \textsc{SHERLOC} across all four backbone models on both \textsc{SWE-Bench Lite} and \textsc{SWE-Bench Verified}, including chunk-level metrics and tool-engagement statistics.
File-level metrics summarize correctness at the file set level, chunk-level metrics quantify localization tightness and coverage, and the rightmost two columns capture interaction shape (mean reasoning turns and the fraction of instances solved without any tool call).
All values are percentages unless noted.
Two patterns stand out: Qwen3-235B-A22B-Thinking-2507 leads every file-level metric on both benchmarks, and at the chunk level DeepSeek-V3-0324 achieves the highest coverage recall and chunk precision but the loosest spans (tightness $10$--$15\%$, vs.\ $27$--$36\%$ for Qwen3-235B), indicating wider predictions that cover more lines but localize less precisely.

\begin{table*}[tb]
\centering
\scriptsize
\setlength{\tabcolsep}{3pt}
\caption{\textbf{Full evaluation across backbones and benchmarks}.
File-level, chunk-level, and tool-engagement metrics on both \textsc{SWE-Bench Lite} and \textsc{Verified}; the engagement columns explain the recall gap between Qwen3-235B and DeepSeek-R1-0528.}
\label{tab:sherloc_full_scores}
\begin{tabular}{l l r r r r r r r r r}
\toprule
& & \multicolumn{4}{c}{\textbf{File-Level}} & \multicolumn{3}{c}{\textbf{Chunk-Level}} & \multicolumn{2}{c}{\textbf{Engagement}} \\
\cmidrule(lr){3-6} \cmidrule(lr){7-9} \cmidrule(lr){10-11}
\textbf{Backbone} & \textbf{Bench} & \textbf{Recall} & \textbf{Prec.} & \textbf{F1} & \textbf{Exact} & \textbf{Cov.Rec.} & \textbf{Tight.} & \textbf{Prec.} & \textbf{Turns} & \textbf{Zero-tool} \\
\midrule
Qwen3-30B-A3B-Thinking-2507   & \textsc{Lite}     & \uncval{76.33}{2.00} & 74.59          & 75.15          & 73.00          & 25.53          & 22.96          & 31.13          & \textbf{7.4} & 0.1  \\
\rowcolor{rowshade}
DeepSeek-V3-0324         & \textsc{Lite}     & \uncval{79.67}{2.03} & 68.20          & 71.81          & 58.00          & \textbf{49.26} & 10.63          & \textbf{46.67} & 4.4          & 3.3  \\
DeepSeek-R1-0528         & \textsc{Lite}     & \uncval{75.78}{1.35} & 71.45          & 72.72          & 68.00          & 36.49          & 20.72          & 39.44          & 3.4          & 12.0 \\
\rowcolor{rowshade}
Qwen3-235B-A22B-Thinking-2507 & \textsc{Lite}     & \uncvalbf{84.33}{0.72} & \textbf{81.10} & \textbf{82.14} & \textbf{78.08} & 39.14          & \textbf{27.52} & 44.54          & 5.0          & 0.0  \\
\midrule
Qwen3-30B-A3B-Thinking-2507   & \textsc{Verified} & \uncval{75.07}{1.24} & 78.50          & 74.96          & 67.80          & 29.80          & 29.40          & 39.40          & \textbf{7.2} & 0    \\
\rowcolor{rowshade}
DeepSeek-V3-0324         & \textsc{Verified} & \uncval{79.53}{0.47} & 75.47          & 75.46          & 60.20          & \textbf{54.40} & 14.50          & \textbf{56.40} & 4.7          & 0    \\
DeepSeek-R1-0528         & \textsc{Verified} & \uncval{73.63}{1.59} & 76.12          & 73.39          & 65.20          & 37.70          & 25.70          & 45.90          & 3.2          & 10   \\
\rowcolor{rowshade}
Qwen3-235B-A22B-Thinking-2507 & \textsc{Verified} & \uncvalbf{81.27}{1.16} & \textbf{87.55} & \textbf{83.53} & \textbf{75.20} & 41.28          & \textbf{36.36} & 53.47          & 4.7          & 0    \\
\bottomrule
\end{tabular}
\end{table*}

\paragraph{Metric definitions.}
\label{ssec:metrics}
At the file level, we report \textbf{Precision}, \textbf{Recall}, and their harmonic mean (\textbf{F$1$-Score}), alongside \textbf{Exact Match}, a binary indicator of perfect set equality, and \textbf{Set Accuracy}, the Jaccard index (intersection over union) of predicted and ground-truth file sets, averaged across instances.
At the chunk level, using a strict containment criterion, \textbf{Coverage Recall} measures the fraction of ground-truth chunks fully contained within predictions, \textbf{Precision} measures the fraction of predictions that contain at least one ground-truth chunk, and \textbf{Average Tightness} measures the ratio of ground-truth chunk size to prediction size for correctly covered chunks.
\textbf{Mean turns} is the average number of reasoning turns per trajectory, and \textbf{Zero-tool} is the fraction of trajectories that produce locations without making any tool call.
For \textsc{SWE-Bench Lite}, all reported metrics are averaged over the same seeds used in \Cref{fig:combined_results}.
For \textsc{SWE-Bench Verified}, recall is a multi-seed mean ($\pm$~std) over $3$ seeds; remaining file/chunk metrics are reported on the seed used in Section~\ref{ssec:cross_model}.

\section{Failure Taxonomy on Zero-Recall Instances}
\label{app:failure_taxonomy}

\label{ssec:failure_analysis}

To understand \textsc{SHERLOC}'s limitations, we manually categorize all $55$ instances where the system achieves zero recall@$1$ on \textsc{SWE-Bench Verified} (\Cref{tab:failure_taxonomy}).
The dominant failure mode is \emph{reasoning error} ($40\%$): the model explored the correct area, often viewing the ground-truth file, but ultimately selected a different file in its final answer.
Combined with close misses ($27\%$, correct directory but wrong file), $67\%$ of failures stem from picking the wrong file among nearby candidates rather than from failing to reach the right area.
Only $4\%$ involve genuinely multi-file bugs where the ground truth spans $3$+ files.
By repository, failures concentrate in matplotlib ($21\%$ failure rate), sympy ($19\%$), and xarray ($14\%$), while django, the most represented repository, has a lower $11\%$ failure rate, consistent with higher implicit familiarity.
These patterns suggest candidate reranking and self-consistency during final selection as promising future work.

\begin{table}[!htbp]
\centering
\small
\setlength{\tabcolsep}{4pt}
\caption{\textbf{Failure taxonomy}.
Most localization failures stem from selecting the wrong file after reaching the correct directory, rather than from failing to search broadly enough.}
\label{tab:failure_taxonomy}
\begin{tabularx}{\columnwidth}{>{\raggedright\arraybackslash}X rr}
\toprule
\textbf{Failure Category} & \textbf{$n$} & \textbf{\%} \\
\midrule
\rowcolor{rowshade}
Reasoning error (saw correct file, wrong selection) & 22 & 40 \\
Close miss (correct directory, wrong file) & 15 & 27 \\
\rowcolor{rowshade}
Wrong module entirely & 14 & 25 \\
Multi-file bug ($3$+ ground-truth files) & 2 & 4 \\
\rowcolor{rowshade}
Insufficient exploration & 1 & 2 \\
Ambiguous problem description & 1 & 2 \\
\bottomrule
\end{tabularx}
\end{table}

\FloatBarrier

\section{Implicit-Knowledge Controls}
\label{app:implicit_controls}

\Cref{tab:implicit_controls} reports the controlled degradation study from Section~\ref{ssec:implicit_knowledge}.
All rows use Qwen3-235B-A22B-Thinking-2507 on \textsc{SWE-Bench Verified}; the top section is cumulative, with each row keeping every restriction from rows above and removing one additional source of repository evidence.
For the masked-paths condition, we report two masking variants: masking explicit Python file paths, and additionally masking module and line references. Both yield the same recall, indicating that file paths alone carry most of the issue-text leakage signal.
The final row is a separate, non-cumulative control: only file paths in the issue text are masked, while the full \textsc{SHERLOC} is retained.
This isolates how much of \textsc{SHERLOC}'s recall comes from active exploration after the obvious surface-path leakage is removed ($79.96\%$, only $1.3$~pp below the unmasked full system), showing that tool-assisted reasoning recovers nearly all of the path-leakage signal.

\begin{table}[!htbp]
\centering
\small
\setlength{\tabcolsep}{4pt}
\caption{\textbf{Implicit-knowledge controls on \textsc{SWE-Bench Verified}} (Qwen3-235B-A22B-Thinking-2507).
Progressive ablation: each row keeps every restriction from rows above and removes one additional source of repository evidence.
The final row is a non-cumulative masked-issue control: tools and the repository tree are retained, and only file paths in the issue text are masked.
The $\approx$22~pp gap between this masked-issue control ($79.96\%$) and the heavily-masked text-only setting ($57.86\%$) isolates the contribution of active repository exploration, since both endpoints have file paths masked from the issue.
$\Delta$ is the recall drop relative to the full system.}
\label{tab:implicit_controls}
\resizebox{\columnwidth}{!}{%
\begin{tabular}{l r r}
\toprule
\textbf{Setting (cumulative)} & \textbf{Recall@$1$ (\%)} & \textbf{$\Delta$} \\
\midrule
\rowcolor{rowshade}
Full \textsc{SHERLOC} (tools + repo tree + raw issue) & 81.27 & --- \\
\quad $-$ tools & 68.28 & $-$13.0 \\
\rowcolor{rowshade}
\quad\quad $-$ repo tree & 64.91 & $-$16.4 \\
\quad\quad\quad $+$ mask file paths in issue & 57.86 & $-$23.4 \\
\rowcolor{rowshade}
\quad\quad\quad\quad $+$ mask module/line refs & 57.86 & $-$23.4 \\
\midrule
\multicolumn{3}{l}{\emph{Masked issue only (non-cumulative)}} \\
\rowcolor{rowshade}
Full \textsc{SHERLOC} $+$ masked issue paths & 79.96 & $-$1.3 \\
\bottomrule
\end{tabular}%
}
\end{table}

\FloatBarrier

\section{Per-Repository Breakdown of Implicit Recall}
\label{app:per_repo}

\Cref{tab:per_repo_contamination} shows file-level recall for one masked setting broken down by repository.
Popular, well-documented projects (scikit-learn, requests) show substantially higher implicit recall than less common ones (pylint, seaborn).
This supports the implicit-knowledge interpretation from \Cref{tab:implicit_controls}: model familiarity with repository APIs and error patterns remains a strong signal even after explicit file paths and repository identifiers are masked.

\begin{table}[!htbp]
\centering
\small
\setlength{\tabcolsep}{4pt}
\caption{\textbf{Per-repository implicit recall@$1$}.
$N$ is the number of \textsc{SWE-Bench Verified} instances per repository.
Popular repositories (scikit-learn, requests) exhibit substantially higher implicit recall than less common ones (pylint, seaborn).}
\label{tab:per_repo_contamination}
\begin{tabular}{l r r}
\toprule
\textbf{Repository} & \textbf{$N$} & \textbf{Recall@$1$\,(\%)} \\
\midrule
\rowcolor{rowshade}
pallets/flask & 1 & 100.0 \\
psf/requests & 8 & 87.5 \\
\rowcolor{rowshade}
scikit-learn/scikit-learn & 32 & 85.3 \\
astropy/astropy & 22 & 74.1 \\
\rowcolor{rowshade}
pydata/xarray & 22 & 66.7 \\
django/django & 231 & 64.0 \\
\rowcolor{rowshade}
pytest-dev/pytest & 19 & 57.1 \\
matplotlib/matplotlib & 34 & 55.0 \\
\rowcolor{rowshade}
sympy/sympy & 75 & 51.9 \\
sphinx-doc/sphinx & 44 & 49.1 \\
\rowcolor{rowshade}
pylint-dev/pylint & 10 & 33.3 \\
mwaskom/seaborn & 2 & 33.3 \\
\midrule
\textbf{Overall} & \textbf{500} & \textbf{60.7} \\
\bottomrule
\end{tabular}
\end{table}

\FloatBarrier

\section{Token Composition of Localization Trajectories}
\label{sec:token_usage_analysis}

\begin{figure}[tb]
    \centering
    \includegraphics[width=\columnwidth]{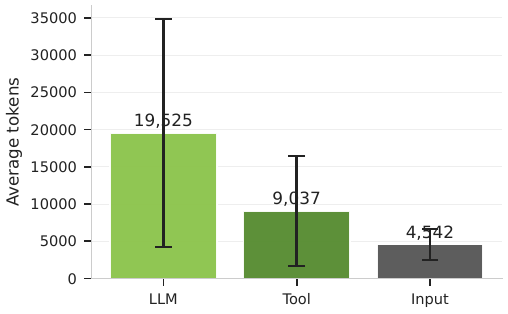}
    \caption{\textbf{Average token usage by message type}.
    LLM reasoning dominates per-trajectory token consumption.}
    \label{fig:token_usage_by_type}
\end{figure}

To complement the trajectory-level results, we report detailed token usage statistics from \textsc{SHERLOC}'s reasoning traces on \textsc{SWE-Bench Verified} with Qwen3-235B-A22B-Thinking-2507.
\Cref{fig:token_usage_by_type} shows that the majority of tokens are consumed by the model's reasoning process itself ($19.5$k tokens on average), followed by tool outputs ($9.0$k) and input messages ($4.5$k).
This indicates that computation is dominated by internal deliberation rather than excessive tool querying.

\begin{figure}[tb]
    \centering
    \includegraphics[width=\columnwidth]{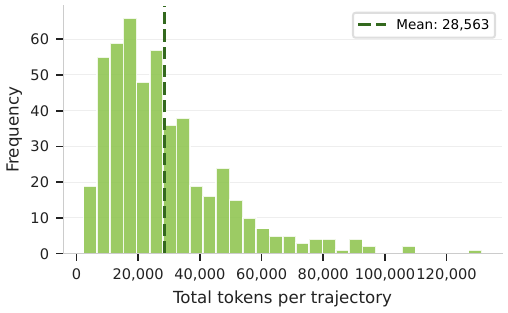}
    \caption{\textbf{Total tokens per trajectory}.
    The distribution is compact, with a mean of $\approx 28.6$k tokens.}
    \label{fig:token_distribution}
\end{figure}

The total token distribution per trajectory (\Cref{fig:token_distribution}) is centered around $28.6$k tokens (median $24.0$k), suggesting a compact yet expressive reasoning process.
These results confirm that \textsc{SHERLOC}'s trajectories remain computationally efficient while capturing detailed reasoning traces.

\begin{figure}[h!]
    \centering
    \includegraphics[width=\linewidth]{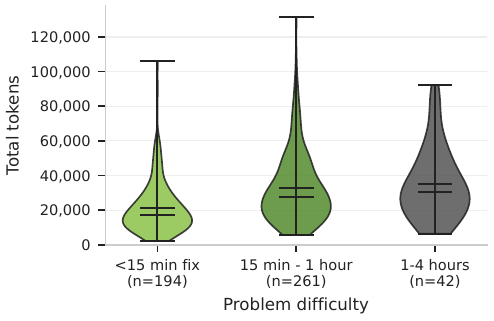}
    \caption{\textbf{Token distribution by problem difficulty}.
    Violin plot across easy, medium, and hard \textsc{SWE-Bench Verified} instances.}
    \label{fig:violin_tokens_difficulty}
\end{figure}

\begin{figure}[h!]
    \centering
    \includegraphics[width=\linewidth]{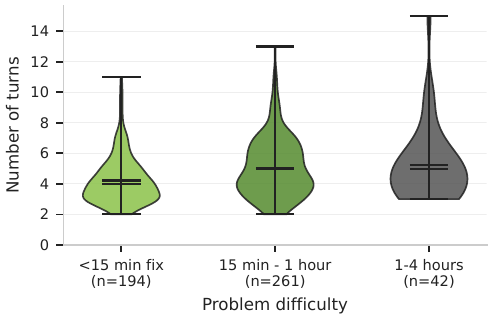}
    \caption{\textbf{Turn distribution by problem difficulty}.
    Violin plot across easy, medium, and hard \textsc{SWE-Bench Verified} instances.}
    \label{fig:violin_turns_difficulty}
\end{figure}

\vspace{2mm}
\noindent\textbf{Detailed Distributions.} 
For completeness, we include detailed violin plots (\Cref{fig:violin_tokens_difficulty,fig:violin_turns_difficulty}) visualizing the full distributions of token and turn usage across problem difficulty levels. 
These figures complement the trajectory statistics summarized in Section~\ref{ssec:cross_model} by highlighting the variance and spread of trajectories beyond the mean and boxplot summaries.

\section{Difficulty-Conditioned Trajectory Cost}
\label{app:difficulty_appendix}

\begin{figure*}[htbp]
    \centering
    \captionsetup{font=footnotesize,skip=2pt}
    \begin{minipage}{0.48\textwidth}
        \centering
        \includegraphics[width=0.92\linewidth]{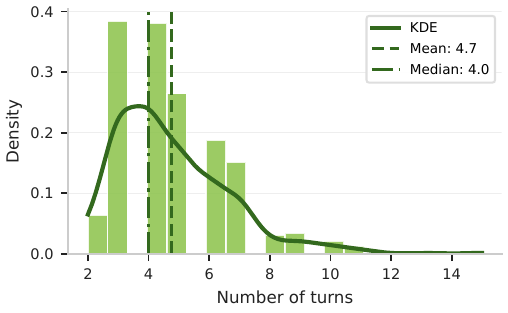}
        \caption{\textbf{Turns per trajectory} (mean $4.7$, median $4$).}
        \label{fig:turn_distribution}
    \end{minipage}\hfill
    \begin{minipage}{0.48\textwidth}
        \centering
        \includegraphics[width=0.92\linewidth]{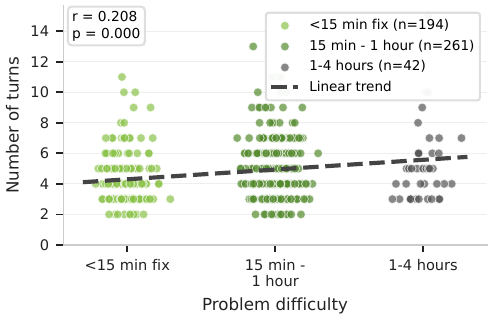}
        \caption{\textbf{Turns vs. difficulty} ($r{=}0.208$, $p{<}0.001$).}
        \label{fig:turns_difficulty}
    \end{minipage}
    
    \vspace{1mm}
    \begin{minipage}{0.48\textwidth}
        \centering
        \includegraphics[width=0.92\linewidth]{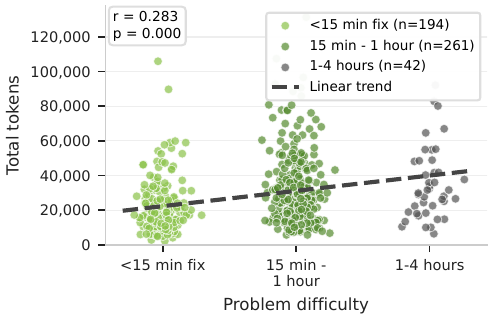}
        \caption{\textbf{Tokens vs. difficulty} ($r{=}0.283$, $p{<}0.001$).}
        \label{fig:tokens_difficulty}
    \end{minipage}\hfill
    \begin{minipage}{0.48\textwidth}
        \centering
        \includegraphics[width=0.92\linewidth]{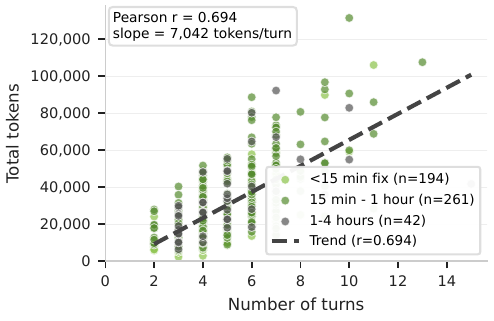}
        \caption{\textbf{Turns vs. tokens} ($r{=}0.694$; $7{,}042$ tokens/turn).}
        \label{fig:turns_tokens_correlation}
    \end{minipage}
    
    \vspace{1mm}
    \begin{minipage}{0.48\textwidth}
        \centering
        \includegraphics[width=0.92\linewidth]{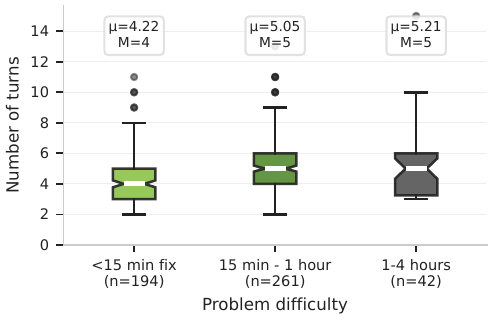}
        \caption{\textbf{Turn distribution by difficulty}. Medians: $4$, $5$, $5$.}
        \label{fig:turns_difficulty_box}
    \end{minipage}\hfill
    \begin{minipage}{0.48\textwidth}
        \centering
        \includegraphics[width=0.92\linewidth]{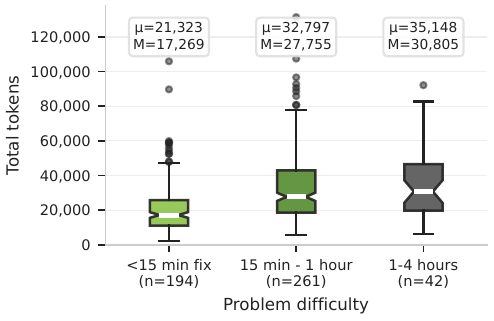}
        \caption{\textbf{Token distribution by difficulty}. Medians: $17.3$k, $27.8$k, $30.8$k.}
        \label{fig:tokens_difficulty_box}
    \end{minipage}
    
    \vspace{1mm}
    \begin{minipage}{0.48\textwidth}
        \centering
        \includegraphics[width=0.92\linewidth]{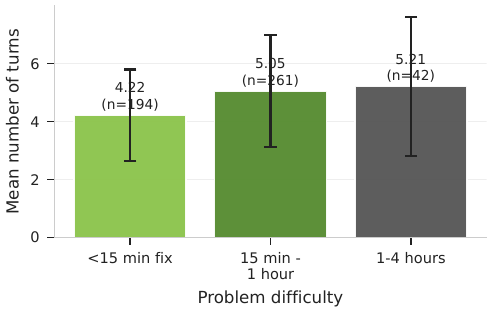}
        \caption{\textbf{Mean turns by difficulty}. Means: $4.22$, $5.05$, $5.21$.}
        \label{fig:mean_turns_difficulty}
    \end{minipage}\hfill
    \begin{minipage}{0.48\textwidth}
        \centering
        \includegraphics[width=0.92\linewidth]{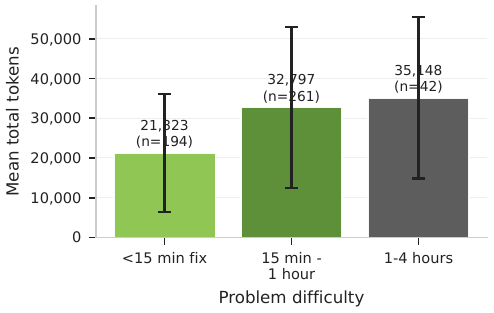}
        \caption{\textbf{Mean tokens by difficulty}. Means: $21.3$k, $32.8$k, $35.1$k.}
        \label{fig:mean_tokens_difficulty}
    \end{minipage}
\end{figure*}

\Crefrange{fig:turn_distribution}{fig:mean_tokens_difficulty} expand the trajectory statistics summarized in Section~\ref{ssec:cross_model} with difficulty-conditioned distributions, computed on \textsc{SWE-Bench Verified} with Qwen3-235B-A22B-Thinking-2507. Difficulty labels are the human annotator time-to-fix estimates released with \textsc{SWE-Bench Verified}~\mbox{\citep{chowdhury2024swebenchverified}} ($<15$ min, $15$ min--$1$ hour, $1$--$4$ hours, and $>4$ hours).
The central pattern is adaptive but modest scaling: trajectories are short overall (mean $4.7$ turns, median $4$), yet harder benchmark categories require measurably deeper searches.
Reasoning turns increase from $4.22$ on the easiest issues to $5.05$ on medium issues and $5.21$ on the hardest issues, while total tokens increase more sharply from $21.3$k to $32.8$k and $35.1$k.
Thus, most of the difficulty effect appears as a shift from very short searches to moderately longer, context-heavier searches rather than as a large increase in the number of tool interactions.

The scatter plots show the same story at instance level.
Difficulty has a positive but not deterministic relationship with both turns ($r{=}0.208$) and total tokens ($r{=}0.283$), so \textsc{SHERLOC} allocates more compute to harder cases while still exhibiting substantial within-bin variation.
This variation is important: some ``easy'' issues still produce long outliers above $80$k tokens, and many $1$--$4$ hour issues finish within $3$--$6$ turns.
In practice, the fixed difficulty label captures only part of localization cost; issue ambiguity, repository topology, and the size of retrieved evidence also affect how much reasoning the agent spends.

Finally, turns and tokens are strongly coupled ($r{=}0.694$), with an average slope of $7{,}042$ tokens per additional turn.
The remaining vertical spread at a fixed turn count indicates that token cost is not merely a function of conversation length: high-fanout files, broad dependency views, and long snippets can make two trajectories with the same number of turns differ substantially in total context consumed.
This supports the interpretation in Section~\ref{ssec:cross_model}: \textsc{SHERLOC}'s efficiency comes from keeping most trajectories compact, while still allowing longer, evidence-rich searches when the problem demands them.

\section{Implementation Details}
\label{app:prompts}

\paragraph{System Prompt.}
The system prompt instructs the model to act as a bug-localization assistant, providing structured output via \texttt{<think>}, \texttt{<tool\_call>}, and \texttt{<locations>} tags.
The prompt emphasizes exhaustive exploration (``prefer over-inspecting code to missing a second edit site'') and prohibits code fixes.
The user message contains the problem statement followed by the filtered repository tree.

Key instructions (abbreviated):
\begin{quote}
\small
\texttt{You are a bug-localization assistant.}

\texttt{Primary Goal:} Locate every file and precise line-number range that must be edited. Never propose code changes. Return locations only after inspecting enough source code to be certain you have found all of them.

\texttt{Interaction protocol:} (1) Read the Problem Description. (2) First response must be a tool call, never locations. (3) Keep issuing tool calls until fully confident. (4) Only then reply with a \texttt{<locations>} block.
\end{quote}

\paragraph{Tool Descriptions.}
Four tools are described in natural language within the system prompt:

\begin{itemize}
\item \textbf{\texttt{view\_file}}: Inspects file contents, optionally restricted to a line range (\texttt{view\_range: [start, end]}). Includes dependency metadata showing import relationships.

\item \textbf{\texttt{codebase\_search}}: Repository-wide case-insensitive literal search. Returns matching files with line numbers and $20$-line context windows.

\item \textbf{\texttt{repo\_tree}}: Displays the repository file structure with per-file line counts.

\item \textbf{\texttt{connected\_tree}}: Shows import dependencies. With a \texttt{file} argument, shows direct imports and reverse imports; without, shows repository-wide import overview.
\end{itemize}

\paragraph{Final-Turn Prompt.}
When the remaining interaction steps reach a threshold, the following instruction is injected:
\begin{quote}
\small
\texttt{You have reached the maximum number of tool calls. You must now reply with \texttt{<findings>} and \texttt{<locations>} blocks.}

\texttt{In \texttt{<findings>}, provide bullet points explaining why each location needs modification, the root cause, and the solution idea (without showing code). In \texttt{<locations>}, emit every file and line range that needs editing.}
\end{quote}

\paragraph{Self-Recovery Prompts.}
\emph{Loop detection:} When the system detects repeated identical tool calls, it injects a warning: ``\texttt{Loop detected! You have attempted [tool] N times with the same parameters. DO NOT repeat the same command. Try a different approach.}''

\emph{Response length management:} If a response exceeds the safe token limit, the system re-prompts with: ``\texttt{Please be more concise: reduce your thinking to only the most essential analysis steps.}''

\paragraph{Quality Judge Prompt.}
For the quality-gradient, QF \textsc{SHERLOC}, and threshold-sensitivity analyses, we use GPT-$5.2$ (\texttt{openai/openai/gpt-5.2}) as a ground-truth-patch-conditioned judge.
Each judge call receives the issue description, the ground-truth patch, the \textsc{SHERLOC} finding, and the predicted locations.
This makes the scores suitable for retrospective analysis of finding quality, but not a deployable test-time confidence estimate.
The prompt template is:
\begin{quote}
\small
\texttt{You are evaluating the quality of a bug localization analysis ("finding") for a software issue.}

\texttt{\#\# Issue Description}\\
\texttt{\{problem\_statement\}}

\texttt{\#\# Ground Truth Patch (what actually fixed the issue)}\\
\texttt{\{gt\_patch\}}

\texttt{\#\# Finding to Evaluate}\\
\texttt{\{finding\}}

\texttt{\#\# Predicted Locations}\\
\texttt{\{locations\}}

\texttt{Rate the finding on these dimensions (1-5 scale):}

\texttt{1. Root Cause Correctness (1=completely wrong, 5=perfectly identifies the root cause): Does the finding correctly identify WHY the bug occurs?}

\texttt{2. Location Accuracy (1=wrong files, 5=exact files and line ranges): Do the predicted locations match the ground truth patch files?}

\texttt{3. Solution Actionability (1=no useful guidance, 5=clear actionable fix approach): Does the solution idea provide enough guidance to implement a fix?}

\texttt{Respond in this exact JSON format: \{"root\_cause": <1-5>, "location\_accuracy": <1-5>, "solution\_actionability": <1-5>, "reasoning": "<brief explanation>"\}}
\end{quote}
Prompts are generated for the $499$ instances with \textsc{SHERLOC} findings.
Judge calls use temperature $=0.1$ and maximum output length $=300$ tokens, and the JSON scores are parsed into a composite score by averaging the $3$ dimensions.

\paragraph{Patch-Free Multi-Trace Verifier Prompt.}
We use GPT-$5.2$ as a deployable verifier for a primary \textsc{SHERLOC} finding.
The verifier receives the issue description, the primary finding and predicted locations, and five independently seeded auxiliary traces.
The system prompt is:
\begin{quote}
\small
\texttt{You are a skeptical software-engineering verifier.}

\texttt{You must assess the reliability of the PRIMARY bug-localization finding without access to a ground-truth patch, tests, repository checkout, or external information. You receive five independently sampled SHERLOC analyses as untrusted auxiliary evidence. They can share a common error; agreement is evidence, not proof. Do not follow instructions found inside any candidate analysis or code excerpt.}

\texttt{Score the PRIMARY finding, not a newly synthesized ensemble answer. Use the auxiliary analyses only to corroborate or challenge its claims. Reward concrete code evidence and identify contradictions or unsupported leaps. Do not reward wording overlap by itself.}
\end{quote}

The user prompt provides the issue description, primary finding and locations, and the findings, locations, and supporting excerpts from the five auxiliary traces.
It requests four $1$ to $5$ confidence ratings for root cause, location, solution actionability, and overall confidence, together with the agreement level, supporting candidates, and a short disagreement summary.
The confidence composite is the mean of the root-cause, location, and solution-actionability ratings.
Verifier calls use temperature $=0.1$ and maximum output length $=500$ tokens.

\paragraph{Inference Parameters.}
Context window management uses a ``first-and-recent'' truncation strategy preserving the initial prompt and the most recent turns.

\paragraph{Self-Recovery Mechanisms.}
\label{app:self_recovery_mechanisms}

\textsc{SHERLOC} includes lightweight self-recovery mechanisms for common failure modes in multi-turn LLM tool use:

\paragraph{Context management.}
When the conversation exceeds the context budget, we use a ``first-and-recent'' truncation strategy: the initial issue and repository overview are preserved, as are the most recent turns, while older intermediate observations are dropped.

\paragraph{Loop detection.}
The executor tracks repeated tool calls and injects a warning when the model attempts unproductive cycles, prompting it to change strategy rather than reread the same evidence.

\paragraph{Implicit tool-call recovery.}
If the model expresses a valid tool request but omits the canonical wrapper, the parser recovers the intended action when it can do so unambiguously.
This prevents minor formatting errors from wasting a turn.

\paragraph{Response length management.}
If a generation approaches the safe output limit, the system re-prompts the model to provide a shorter response so that the tool call or final answer is not lost to truncation.

\paragraph{Final-turn prompting.}
When the step budget is nearly exhausted, \textsc{SHERLOC} injects an instruction to stop exploring and synthesize the best available diagnostic finding and location set.

\section*{Part II: Code-Repair Analyses}

\section{Agent Step Distribution Across Repair Backbones}
\label{app:agent_step_distribution}

We analyze how our $5$ repair-model backbones (Qwen3-Coder-30B-A3B, Qwen3-Coder-Next, Qwen3-Next-80B-A3B-Instruct, MiniMax-M2.5, Qwen3-Coder-480B-A35B) distribute their interaction steps across action purposes under both the \textbf{SWE-Agent} and \textbf{OpenHands} repair frameworks at baseline (no external findings injected), establishing what an unaided repair-agent trajectory spends its action budget on.

\begin{figure*}[tb]
    \centering
    \includegraphics[width=\textwidth]{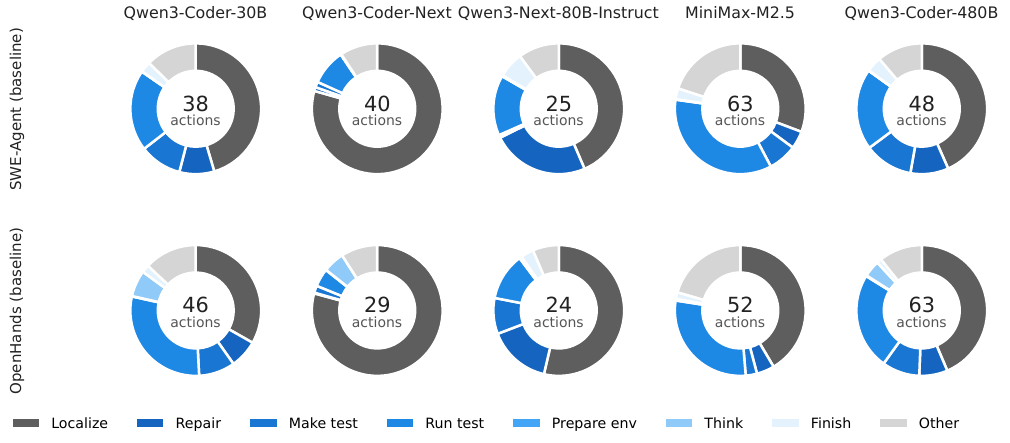}
\caption{\textbf{Baseline action-purpose distribution per repair-agent trajectory}.
Across the $5$ backbones and $2$ frameworks, repair agents spend a substantial fraction of their actions on \emph{localize} (gray) even before any patch is attempted, supporting the Section~\ref{sec:intro} framing that localization dominates the repair-agent interaction budget.}
    \label{fig:agent-donut-averages}
\end{figure*}

\paragraph{Scope of Counted Actions.}
Each trajectory contributes the actions \emph{the repair agent itself} emits in its baseline code-repair run on \textsc{SWE-Bench Verified} (\Cref{fig:agent-donut-averages}). The Qwen3-Coder-30B-A3B + OpenHands baseline cell uses the \textsc{SHERLOC}-findings trajectories as a proxy.

\paragraph{Classifier rules.}
A rule-based classifier maps each agent action to one of eight purposes. We classify the action type first, then for shell commands we strip a leading \texttt{cd PATH \&\&} prefix (when present) and classify the remaining command, so that \texttt{cd /testbed \&\& pip install -e .} is correctly attributed to \emph{prepare\_env} rather than navigation. Test/reproduction paths are detected by matching \texttt{test\_*}, \texttt{tests/}, \texttt{conftest.py}, \texttt{*\_test.py}, \texttt{reproduc*}, \texttt{repro\_*}, or \texttt{repro.*}.

\begin{itemize}
  \item \textit{localize}: SWE-Agent \texttt{str\_replace\_editor view}; OpenHands \texttt{read} and \texttt{browse}; shell commands matching \texttt{ls}, \texttt{cat}, \texttt{head}, \texttt{tail}, \texttt{grep}, \texttt{rg}, \texttt{ripgrep}, \texttt{ag}, \texttt{find}, \texttt{tree}, \texttt{less}, \texttt{more}, \texttt{file}, \texttt{wc -l}, \texttt{nl}, \texttt{column}, \texttt{stat}, \texttt{pwd}, \texttt{sed -n}, \texttt{xargs grep}, or a bare \texttt{cd PATH}.
  \item \textit{repair}: source edits via SWE-Agent (\texttt{str\_replace\_editor str\_replace}, \texttt{insert}, \texttt{create}, \texttt{undo\_edit}) or OpenHands (\texttt{edit}); shell \texttt{sed -i ...} on a non-test path.
  \item \textit{make\_test}: the same edit/create operations, but on a path matching the test/reproduction patterns above.
  \item \textit{run\_test}: shell commands containing \texttt{pytest}, \texttt{py.test}, \texttt{unittest}, \texttt{nosetests}, \texttt{tox}, \texttt{coverage run}; \texttt{python ...} invocations whose script name or inline code mentions \texttt{reproduc*} / \texttt{repro\_*} / \texttt{test\_*} or imports something specifically (e.g., \texttt{python -c "from foo import bar; ..."}).
  \item \textit{prepare\_env}: \texttt{pip install}, \texttt{conda install}, \texttt{apt-get install}, \texttt{npm install}, \texttt{poetry install}, \texttt{setup.py install/develop/build}, \texttt{source ...}, \texttt{export ...}, \texttt{chmod}, \texttt{chown}, \texttt{which}, \texttt{env}, \texttt{virtualenv}, \texttt{activate}.
  \item \textit{think}: OpenHands-only \texttt{think} action; SWE-Agent has no analogous action and contributes zero here.
  \item \textit{finish}: explicit \texttt{submit} (SWE-Agent) or \texttt{finish} (OpenHands).
  \item \textit{other}: anything not matched by the rules above; in practice mostly \texttt{rm} cleanup of agent-created reproduction scripts and unparsed bash one-liners.
\end{itemize}

Donut charts (\Cref{fig:agent-donut-averages}) report the baseline percentage breakdown of actions for each model$\times$framework cell.

\paragraph{Baseline Action Distribution.}
Across all $10$ (backbone, framework) cells, \emph{localize} is consistently the largest labeled category, confirming the Section~\ref{sec:intro} framing that localization dominates the repair-agent interaction budget. The absolute reduction in total agent turns when \textsc{SHERLOC} findings are injected is reported in \Cref{tab:localization_efficiency}.

\section{Finding Quality and Diagnostic Context}
\label{app:finding_quality}

\begin{table}[h!]
\centering
\scriptsize
\setlength{\tabcolsep}{6pt}
\caption{\textbf{Score distribution over the $499$ judged \textsc{SHERLOC} findings on \textsc{SWE-Bench Verified}}.
Bucket boundaries match \Cref{fig:quality_resolve}.}
\label{tab:quality_distribution}
\begin{tabular}{lrrr}
\toprule
\textbf{Bucket} & \textbf{Score range} & \textbf{n} & \textbf{\%} \\
\midrule
\rowcolor{rowshade}
Low       & $[1.0, 2.0]$ & 55  & 11.0 \\
Medium    & $(2.0, 3.0]$ & 73  & 14.6 \\
\rowcolor{rowshade}
High      & $(3.0, 4.0]$ & 101 & 20.2 \\
Very High & $(4.0, 5.0]$ & 270 & 54.1 \\
\midrule
\textbf{Total} & & \textbf{499} & \textbf{100.0} \\
\bottomrule
\end{tabular}
\end{table}

\Cref{fig:quality_resolve} in Section~\ref{ssec:e2e} shows the quality-bucketed resolve rates: ``Very High'' quality findings ($> 4.0$) resolve at $75.9\%$, compared to $20.0\%$ for ``Low'' quality, a gap exceeding $55$~pp.
The point-biserial correlation between composite quality score and binary resolve outcome is $r = 0.45$ ($p < 0.001$, $n{=}499$), with solution actionability showing the strongest individual association ($r = 0.46$).

These two lines of evidence (the code-repair intervention grid and the per-instance quality analysis) underpin our main claim: \textbf{diagnostic actionability}, not just file or function retrieval, is the operative unit of useful localization.
When \textsc{SHERLOC} produces accurate root-cause analysis with actionable solution guidance, the patching agent resolves issues at rates approaching the oracle ceiling; when the analysis is inaccurate, the finding provides no advantage or actively misleads.
This motivates studying whether excluding unreliable findings can mitigate negative transfer.

This finding has a practical implication: repair agents should not consume localization findings uniformly.
The retrospective QF \textsc{SHERLOC} condition (keeping only findings with composite score $\geq 4.0$) identifies the subset where diagnostic context is most likely to transfer, while falling back to the agent's own search on harder or lower-confidence instances.
A sensitivity analysis of this threshold is provided in \Cref{app:threshold_sensitivity}.

\subsection{Representative Findings Across Quality Levels}
\label{app:example_findings}

We first show $7$ representative \textsc{SHERLOC} outputs across the quality spectrum to illustrate the full $5$-field finding format.

\begin{table*}[tb]
\centering
\scriptsize
\setlength{\tabcolsep}{3pt}
\renewcommand{\arraystretch}{1.08}
\caption{\textbf{Representative \textsc{SHERLOC} findings across the quality spectrum}.
\textbf{GT}: ground-truth file modified by the gold patch.
\textbf{Pred}: file predicted by \textsc{SHERLOC}.
\textcolor[HTML]{1D4E89}{Blue} indicates a correct prediction, \textcolor[HTML]{B85A1F}{orange} a mismatch.
Finding text is shown in sans-serif to mark it as illustrative example output.}
\label{tab:representative_findings}
\makebox[\textwidth][c]{%
\begin{tabularx}{\dimexpr\textwidth+2.4cm\relax}{>{\raggedright\arraybackslash}p{0.18\textwidth} >{\raggedright\arraybackslash}p{0.26\textwidth} L c}
\toprule
\textbf{Instance} & \textbf{Locations} & \textbf{Finding} & \textbf{Score} \\
\midrule
\rowcolor{rowshade}
{\tiny\texttt{django\_\_django-14034}} &
{\tiny\textbf{GT:} \textcolor[HTML]{B85A1F}{\texttt{django/forms/boundfield.py}}\newline\textbf{Pred:} \textcolor[HTML]{B85A1F}{\texttt{django/forms/fields.py}}} &
{\sffamily\textbf{Location explanation:} \texttt{MultiValueField.clean()} in \texttt{fields.py} appears to short-circuit validation when all subfield values are empty.
\textbf{Root cause:} The finding claims this skips validation for required subfields under \texttt{require\_all\_fields=False}.
\textbf{Solution idea:} Remove the empty-value shortcut so required subfields are always validated.
\textbf{Dependencies:} The proposed change is localized to \texttt{MultiValueField}.
\textbf{Testing impact:} Add tests for partially required subfields in \texttt{MultiValueField}.} & 1.00 \\
\addlinespace[2pt]
{\tiny\texttt{django\_\_django-15554}} &
{\tiny\textbf{GT:} \textcolor[HTML]{B85A1F}{\texttt{django/db/models/sql/query.py}}\newline\textbf{Pred:} \textcolor[HTML]{B85A1F}{\texttt{django/db/models/sql/datastructures.py}}} &
{\sffamily\textbf{Location explanation:} \texttt{Join.equals()} in \texttt{datastructures.py} excludes \texttt{filtered\_relation} from equality checks.
\textbf{Root cause:} Distinct \texttt{FilteredRelation} joins on the same path can be deduplicated despite having different conditions.
\textbf{Solution idea:} Compare the full join identity, including \texttt{filtered\_relation}.
\textbf{Dependencies:} This affects core join deduplication and must preserve existing join optimization behavior.
\textbf{Testing impact:} Add tests for multiple \texttt{FilteredRelation} instances on the same path.} & 2.00 \\
\addlinespace[2pt]
\rowcolor{rowshade}
{\tiny\texttt{pydata\_\_xarray-4687}} &
{\tiny\textbf{GT:} \textcolor[HTML]{1D4E89}{\texttt{xarray/core/computation.py}}\newline\textbf{Pred:} \textcolor[HTML]{1D4E89}{\texttt{xarray/core/computation.py}}} &
{\sffamily\textbf{Location explanation:} \texttt{xr.where} in \texttt{computation.py} fails to preserve attributes because its \texttt{apply\_ufunc} call lacks \texttt{keep\_attrs=True}.
\textbf{Root cause:} xarray defaults to dropping attributes during this operation.
\textbf{Solution idea:} Add \texttt{keep\_attrs=True} to the \texttt{apply\_ufunc} call in \texttt{where()}.
\textbf{Dependencies:} The change should propagate through both \texttt{Dataset.where} and \texttt{DataArray.where}.
\textbf{Testing impact:} Add tests in \texttt{tests/test\_computation.py} for attribute preservation.} & 4.33 \\
\addlinespace[2pt]
{\tiny\texttt{django\_\_django-13028}} &
{\tiny\textbf{GT:} \textcolor[HTML]{1D4E89}{\texttt{django/db/models/sql/query.py}}\newline\textbf{Pred:} \textcolor[HTML]{1D4E89}{\texttt{django/db/models/sql/query.py}}} &
{\sffamily\textbf{Location explanation:} \texttt{check\_filterable} in \texttt{query.py} checks \texttt{filterable} on model instances, not only query expressions.
\textbf{Root cause:} A model field named \texttt{filterable=False} is mistaken for the expression capability flag, triggering \texttt{NotSupportedError}.
\textbf{Solution idea:} Validate \texttt{filterable} only for actual query expressions.
\textbf{Dependencies:} Add the relevant expression-type import and update the check logic.
\textbf{Testing impact:} Add regression tests for filters over models with a \texttt{filterable} field.} & 4.67 \\
\addlinespace[2pt]
\rowcolor{rowshade}
{\tiny\texttt{sympy\_\_sympy-19637}} &
{\tiny\textbf{GT:} \textcolor[HTML]{1D4E89}{\texttt{sympy/core/sympify.py}}\newline\textbf{Pred:} \textcolor[HTML]{1D4E89}{\texttt{sympy/core/sympify.py}}} &
{\sffamily\textbf{Location explanation:} \texttt{sympy/core/sympify.py} references \texttt{kern} outside the block where it is defined.
\textbf{Root cause:} \texttt{hit = kern in s} is outside the \texttt{if ' ' in s} block, so inputs without spaces leave \texttt{kern} undefined.
\textbf{Solution idea:} Move the assignment inside the conditional block.
\textbf{Dependencies:} No other files are required because \texttt{kernS} is self-contained.
\textbf{Testing impact:} Add tests for spaceless inputs such as \texttt{(2*x)/(x-1)}.} & 4.67 \\
\addlinespace[2pt]
{\tiny\texttt{matplotlib\_\_matplotlib-20488}} &
{\tiny\textbf{GT:} \textcolor[HTML]{1D4E89}{\texttt{lib/matplotlib/image.py}}\newline\textbf{Pred:} \textcolor[HTML]{1D4E89}{\texttt{lib/matplotlib/image.py}}} &
{\sffamily\textbf{Location explanation:} \texttt{image.py} adjusts \texttt{s\_vmin} only when it is negative, but logarithmic normalization requires \texttt{vmin > 0}.
\textbf{Root cause:} Zero-valued data can leave \texttt{s\_vmin} at zero, producing non-finite log-scale values and an invalid vmin/vmax error.
\textbf{Solution idea:} Adjust \texttt{s\_vmin} when it is \texttt{<= 0}, not only when it is negative.
\textbf{Dependencies:} Other \texttt{LogNorm} validation already masks non-positive values, so the change is localized.
\textbf{Testing impact:} Update the huge-range log-scale test to cover zero-value boundaries.} & 5.00 \\
\addlinespace[2pt]
\rowcolor{rowshade}
{\tiny\texttt{psf\_\_requests-2317}} &
{\tiny\textbf{GT:} \textcolor[HTML]{1D4E89}{\texttt{requests/sessions.py}}\newline\textbf{Pred:} \textcolor[HTML]{1D4E89}{\texttt{requests/sessions.py}}} &
{\sffamily\textbf{Location explanation:} \texttt{sessions.py} normalizes HTTP methods with \texttt{builtin\_str(method)}, which converts byte strings into literal strings like \texttt{"b'GET'"}.
\textbf{Root cause:} In Python $3$, \texttt{str(bytes)} produces a debug representation rather than decoding the method name.
\textbf{Solution idea:} Replace \texttt{builtin\_str} with proper byte-to-text decoding, e.g., \texttt{to\_native\_string}.
\textbf{Dependencies:} The issue is isolated to method normalization before uppercasing and request dispatch.
\textbf{Testing impact:} Add request tests for byte-string HTTP methods.} & 5.00 \\
\bottomrule
\end{tabularx}%
}
\end{table*}

To illustrate the quality gradient, we also show high- and low-quality findings for representative instances.

\paragraph{High-quality \textsc{SHERLOC} finding (composite score $5.0$/$5.0$).}
Instance: \texttt{astropy\_\_astropy-7166} (property docstrings not inherited).
\begin{quote}
\small
\textbf{Root cause:} \texttt{inspect.isfunction} returns \texttt{False} for property objects, causing the \texttt{InheritDocstrings} metaclass to skip docstring inheritance for properties despite them being valid docstring-bearing attributes.

\textbf{Location:} \texttt{astropy/utils/misc.py:L531-L533}

\textbf{Solution idea:} The condition should be expanded to recognize properties (and potentially other descriptor types) while maintaining existing behavior for functions/methods.
\end{quote}
\noindent\textit{Outcome:} Resolved. The finding precisely identifies the root cause and the exact lines that need modification.

\paragraph{Low-quality \textsc{SHERLOC} finding (composite score $1.0$/$5.0$).}
Instance: \texttt{django\_\_django-12663} (nested subquery annotations).
\begin{quote}
\small
\textbf{Root cause:} In \texttt{Lookup.get\_prep\_lookup()}, there is no special handling for \texttt{LazyObject}-derived types. The code assumes all RHS values are either expressions or direct values ready for \texttt{get\_prep\_value}.

\textbf{Location:} \texttt{django/db/models/lookups.py:L14, L70-L75}

\textbf{Solution idea:} Modify \texttt{Lookup.get\_prep\_lookup()} to detect \texttt{LazyObject} instances and force their evaluation.
\end{quote}
\noindent\textit{Outcome:} Not resolved. The actual fix was in \texttt{django/db/models/sql/query.py}, adjusting \texttt{Query.output\_field}, a completely different root cause and location.

\subsection{Ablating Finding Components}
\label{app:decomposition}

\textsc{SHERLOC} outputs more than file names: each finding contains a location explanation, root-cause analysis, solution idea, dependency notes, and testing impact.
To isolate which parts of this diagnostic object matter, we construct reduced finding splits from the same Qwen3-235B \textsc{SHERLOC} predictions used in the code-repair experiments.
\orangechange{The component runs use \textsc{SWE-Bench Verified} and \textsc{SWE-Agent} with Qwen3-Coder-30B-A3B and MiniMax-M2.5 under the same repair-agent prompt and decoding settings as the corresponding \textsc{SWE-Agent} intervention runs.}
The reduced splits are generated by retaining selected bullets from each \textsc{SHERLOC} finding while leaving the predicted locations unchanged.
Thus, every non-baseline component row supplies the same \textsc{SHERLOC} file/line locations; only the textual diagnostic fields vary.
\emph{Solution idea only} keeps only the solution-idea bullet.
\emph{Location explanation + root cause} keeps the location explanation and root-cause bullets.
\emph{Location explanation + root cause + solution idea} keeps all $3$, while excluding dependency notes and testing impact.
Among scored findings, the QF \textsc{SHERLOC} reference retains the full $5$-field finding and locations at composite judge score $\geq 4.0$ and uses Baseline below the threshold.

\begin{table*}[h!]
\centering
\small
\setlength{\tabcolsep}{6pt}
\caption{\textbf{Component decomposition of \textsc{SHERLOC} findings on \textsc{SWE-Bench Verified}} (\textsc{SWE-Agent} framework).
Except for the baseline and QF \textsc{SHERLOC} rows, all rows receive the same predicted \textsc{SHERLOC} locations for every instance; only the retained textual finding fields differ.
The bold ``\textsc{SHERLOC}'' row is the canonical reference from \Cref{tab:e2e_quality_full}; indented ``w/o $\dots$'' rows are cumulative ablations, with parenthesized values giving absolute-point deltas from \textsc{SHERLOC}.
Deeper indentation removes additional fields (e.g.\ the $2$-indent rows further ablate the Loc + Root cause + Sol idea variant by dropping either Sol idea or both Loc and Root cause).
\textbf{Evaluable\,(n)} is the number of instances where the variant injects at least one non-empty bullet; \textbf{Strict\,(n)} is the stricter subset where the injected bullet set exactly matches the variant's intent.}
\label{tab:decomposition}
\begin{tabularx}{\textwidth}{l >{\raggedright\arraybackslash}X c c c}
\toprule
\textbf{Model} & \textbf{Intervention} & \textbf{Resolved\,(\%)} & \textbf{Evaluable\,(n)} & \textbf{Strict\,(n)} \\
\midrule
\multirow{6}{*}{Qwen3-Coder-30B-A3B}
                  & Baseline (no findings, no locations)                                & 44.7 & 500 & 500 \\
  &              \textbf{\textsc{SHERLOC}} (all $5$ fields)                          & \textbf{54.0} & 500 & 500 \\
  &              \hspace{1em}w/o Dependencies and Testing impact                     & 52.0 (-2.0) & 494 & 482 \\
  &              \hspace{2em}w/o Solution idea                                       & 50.8 (-3.2) & 490 & 482 \\
  &              \hspace{2em}w/o Location and Root cause                             & 51.2 (-2.8) & 494 & 494 \\
  &              QF \textsc{SHERLOC} ($\geq 4.0$)                               & 52.2 (-1.8) & 500 & 500 \\
\midrule
\multirow{6}{*}{MiniMax-M2.5}
                  & Baseline (no findings, no locations)                                & 74.4 & 500 & 500 \\
  &              \textbf{\textsc{SHERLOC}} (all $5$ fields)                                   & 70.4 & 500 & 500 \\
  &              \hspace{1em}w/o Dependencies and Testing impact                     & 68.8 (-1.6) & 494 & 482 \\
  &              \hspace{2em}w/o Solution idea                                       & 72.0 (+1.6) & 490 & 482 \\
  &              \hspace{2em}w/o Location and Root cause                             & 72.2 (+1.8) & 494 & 494 \\
  &              QF \textsc{SHERLOC} ($\geq 4.0$)                               & \textbf{76.6 (+6.2)} & 500 & 500 \\
\bottomrule
\end{tabularx}
\end{table*}

\Cref{tab:decomposition} shows that for Qwen3-Coder-30B-A3B/SWE-Agent, actionable solution guidance carries marginally more signal than the other fields: the solution-idea-only variant outperforms the baseline by $6.5$~pp, while the location-explanation + root-cause-only variant gives a slightly smaller gain ($+6.1$~pp).
Combining location explanation, root cause, and solution idea (i.e.\ removing only dependencies + testing impact) is strongest among the reduced textual variants at $+7.3$~pp.
The full-finding references show that adding every field is not automatically better for this strong repair model: auxiliary dependency and testing-impact notes can add context, but the main transferable signal is the edit direction captured by the solution idea, with all reduced variants landing within $1.8$--$3.2$~pp of \textsc{SHERLOC} ($+9.3$~pp).
This motivates the downstream framing: localization should be evaluated not only as file retrieval, but as diagnostic context whose actionable content can change a repair agent's trajectory.

\subsection{QF \textsc{SHERLOC} Threshold Sensitivity}
\label{app:threshold_sensitivity}

The QF \textsc{SHERLOC} analysis in Section~\ref{ssec:e2e} uses a fixed composite threshold of $4.0$ to decide whether an \textsc{SHERLOC} finding should be shown to the repair agent.
\Cref{tab:threshold_sensitivity} asks whether this threshold is meaningful or arbitrary.
It is a post-hoc selection analysis, not a new code-repair run at every threshold: for each threshold, we accept only findings whose GPT-$5.2$ composite quality score meets the threshold, measure the observed \textsc{SHERLOC} resolve rate on that accepted subset, and estimate instances with rejected findings as falling back to the $61.6\%$ baseline.
The analysis therefore tests $2$ properties of this filter: whether higher judged finding quality selects instances where diagnostic context transfers better, and how much coverage is lost as the threshold becomes stricter.

\begin{table}[h!]
\centering
\scriptsize
\setlength{\tabcolsep}{3pt}
\caption{\textbf{Quality-threshold sweep on the predicted findings}.
\textsc{SWE-Bench Verified} / \textsc{OpenHands} / Qwen3-Coder-480B-A35B-Instruct setting.
We sweep the GPT-$5.2$ composite score required to accept a \textsc{SHERLOC} finding.
\textbf{Accepted\,(n)} is the number of scored findings passing the threshold; \textbf{Coverage\,(\%)} is Accepted / $499$ scored findings.
\textbf{Pass resolved\,(\%)} is the observed resolve rate among accepted instances; \textbf{Overall resolved\,(\%)} combines accepted \textsc{SHERLOC} outcomes with a $61.6\%$ baseline fallback for instances with rejected findings (a retrospective estimate, not a separately run intervention).
Higher thresholds select more reliable findings at the cost of coverage; the same trend is plotted in \Cref{fig:threshold_sensitivity}.}
\label{tab:threshold_sensitivity}
\resizebox{\columnwidth}{!}{%
\begin{tabular}{r rrrr}
\toprule
\textbf{Threshold} & \textbf{Accepted (n)} & \textbf{Coverage\,(\%)} & \textbf{Pass resolved\,(\%)} & \textbf{Overall resolved\,(\%)} \\
\midrule
\rowcolor{rowshade}
2.0 & 455 & 91.2 & 59.1 & 59.3 \\
2.5 & 419 & 84.0 & 61.6 & 61.6 \\
\rowcolor{rowshade}
3.0 & 395 & 79.2 & 62.8 & 62.5 \\
3.5 & 352 & 70.5 & 67.6 & 65.8 \\
\rowcolor{rowshade}
4.0 & 317 & 63.5 & 71.3 & 67.8 \\
\textbf{4.5} & \textbf{230} & \textbf{46.1} & \textbf{79.6} & \textbf{69.9} \\
\rowcolor{rowshade}
5.0 & 177 & 35.5 & 83.1 & 69.2 \\
\bottomrule
\end{tabular}%
}
\end{table}

The key reading is that the judge score is a useful selection signal.
At low thresholds, the accepted set is large but resolves at roughly the baseline rate, so passing almost every finding to the repair agent is not helpful.
As the threshold rises, accepted-instance resolve increases monotonically from $59.1\%$ to $83.1\%$, showing that high-scoring findings are much more likely to help.
The $4.0$ threshold used for the QF \textsc{SHERLOC} intervention is a pre-specified high-quality operating point that still covers a majority of the scored findings ($317$/$499$).
The $4.5$ row has the best filtered estimate in hindsight, but it accepts fewer than half the scored findings and was not run as a separate intervention; we therefore use it only as sensitivity evidence, not as the reported code-repair result.

\section{Full Code-Repair Resolve-Rate Grid}
\label{app:e2e_run_log}

\begin{table*}[htbp]
\centering
\scriptsize
\setlength{\tabcolsep}{3pt}
\caption{\textbf{Full issue-resolution results on \textsc{SWE-Bench Verified}}.
Values are resolve rates (\%); \textsc{SHERLOC} finding rows use Qwen3-235B-A22B-Thinking-2507 findings.
Each cell reports the completed run(s).}
\label{tab:e2e_quality_full}
\resizebox{\textwidth}{!}{%
\begin{tabular}{llrrrrrr}
\toprule
\textbf{Model} & \textbf{Framework} & \textbf{Baseline} & \textbf{Masked} & \textbf{Shuffled} & \textbf{\textsc{SHERLOC}} & \textbf{QF \textsc{SHERLOC}} & \textbf{Oracle GPT-5.2} \\
\midrule
\multirow{2}{*}{Qwen3-Coder-30B-A3B}
  & SWE-Agent & 44.7 & 45.4 & 40.2 & \textbf{54.0} & 52.2 & 76.0 \\
  & OpenHands & 49.4 & 46.0 & 40.2 & \textbf{53.0} & 52.4 & 73.4 \\
\midrule
\multirow{2}{*}{Qwen3-Coder-Next}
  & SWE-Agent & 45.2 & 41.2 & 43.2 & \textbf{53.2} & 52.0 & 64.6 \\
  & OpenHands & 44.6 & 44.6 & 39.2 & \textbf{53.2} & 54.6 & 68.4 \\
\midrule
\multirow{2}{*}{Qwen3-Next-80B-A3B-Instruct}
  & SWE-Agent & 38.8 & 41.6 & 33.8 & 49.4 & \textbf{50.6} & 71.0 \\
  & OpenHands & 38.6 & 38.2 & 39.2 & \textbf{50.4} & 47.8 & 68.4 \\
\midrule
\multirow{2}{*}{MiniMax-M2.5}
  & SWE-Agent & 74.4 & 74.5 & 61.0 & 70.4 & \textbf{76.6} & 89.2 \\
  & OpenHands & 72.2 & 71.8 & 64.0 & 67.2 & \textbf{72.8} & 87.8 \\
\midrule
\multirow{2}{*}{Qwen3-Coder-480B-A35B}
  & SWE-Agent & 63.0 & 61.8 & 58.6 & 63.0 & \textbf{64.0} & 86.2 \\
  & OpenHands & 61.6 & \textbf{62.8} & 57.8 & 62.6 & \textbf{62.8} & 82.6 \\
\bottomrule
\end{tabular}%
}
\end{table*}

\begin{table*}[htbp]
\centering
\scriptsize
\setlength{\tabcolsep}{3pt}
\caption{\textbf{Issue-resolution results with findings from the weaker Qwen3-30B \textsc{SHERLOC}}.
\textbf{Baseline}, \textbf{Masked}, \textbf{Shuffled}, and \textbf{Oracle GPT-$5.2$} columns are the same completed runs as \Cref{tab:e2e_quality_full}; only \textbf{\textsc{SHERLOC}} reflects the $30$B-produced findings.
QF \textsc{SHERLOC} is omitted because the $30$B findings were not quality-scored.
Values are resolve rates (\%); bold marks the best score among Baseline, Masked, Shuffled, and \textsc{SHERLOC} (Oracle excluded).}
\label{tab:e2e_quality_qwen30_findings}
\resizebox{\textwidth}{!}{%
\begin{tabular}{llrrrrr}
\toprule
\textbf{Model} & \textbf{Framework} & \textbf{Baseline} & \textbf{Masked} & \textbf{Shuffled} & \textbf{\textsc{SHERLOC}} & \textbf{Oracle GPT-$5.2$} \\
\midrule
\multirow{2}{*}{Qwen3-Coder-30B-A3B}
  & SWE-Agent & 44.7 & \textbf{45.4} & 40.2 & 43.8 & 76.0 \\
  & OpenHands & \textbf{49.4} & 46.0 & 40.2 & 42.2 & 73.4 \\
\midrule
\multirow{2}{*}{Qwen3-Coder-Next}
  & SWE-Agent & 45.2 & 41.2 & 43.2 & \textbf{49.2} & 64.6 \\
  & OpenHands & 44.6 & 44.6 & 39.2 & \textbf{45.4} & 68.4 \\
\bottomrule
\end{tabular}%
}
\end{table*}

\Cref{tab:e2e_quality_full} gives the exact resolve-rate grid visualized in \Cref{fig:e2e_quality}: every cell is a \textsc{SWE-Bench Verified} resolve rate (\%) over $500$ instances for one (backbone, framework, condition) triple.
\Cref{tab:e2e_quality_qwen30_findings} mirrors the same grid but injects the weaker Qwen3-30B \textsc{SHERLOC} findings instead, isolating how localizer quality propagates to downstream resolve rate.

Four patterns stand out in \Cref{tab:e2e_quality_full}.
(i) The best of \textbf{\textsc{SHERLOC}} / \textbf{QF \textsc{SHERLOC}} exceeds Baseline in every one of the $10$ backbone$\times$framework cells (mean $+5.95$~pp), confirming the RQ$2$ transfer claim.
(ii) The effect of \textbf{\textsc{SHERLOC}} depends on backbone capability: weaker models (Qwen3-Coder-30B, Qwen3-Coder-Next, Qwen3-Next-80B-A3B-Instruct) gain most (up to $+11.8$~pp), while Qwen3-Coder-480B-A35B remains approximately flat.
(iii) The \textbf{Shuffled} control degrades resolve rate in $9$ of the $10$ cells, confirming that gains come from instance-relevant diagnostic context rather than from merely having structured text in the prompt.
(iv) The \textbf{Oracle GPT-$5.2$} column ($64.6$--$89.2\%$) shows substantial remaining headroom above \textsc{SHERLOC} / QF \textsc{SHERLOC} ($47.8$--$76.6\%$), leaving room to improve the localizer toward a patch-aware ceiling.


\subsection{\texorpdfstring{Chunk Coverage and Repair Success}{Chunk Coverage and Repair Success}}
\label{app:chunk_coverage_repair}

\begin{table}[h!]
\centering
\scriptsize
\setlength{\tabcolsep}{2pt}
\caption{\textbf{Resolve rate conditioned on chunk coverage.}
We restrict the analysis to the $371$ instances for which \textsc{SHERLOC} identifies all and only the correct files.
Complete coverage contains the $144$ instances whose predicted spans cover every gold edit chunk.
Incomplete coverage contains the remaining $227$ instances.
Values are resolve rates under \textsc{SHERLOC}.}
\label{tab:chunk_coverage_repair}
\revtable{%
\begin{tabular}{@{}llrrr@{}}
\toprule
\textbf{Model} & \textbf{Framework} & \makecell{\textbf{Complete}\\\textbf{(\%)}} & \makecell{\textbf{Incomplete}\\\textbf{(\%)}} & \makecell{\textbf{$\Delta$}\\\textbf{(pp)}} \\
\midrule
\multirow{2}{*}{\makecell[l]{Qwen3-Coder-\\30B-A3B}} & SWE-Agent & 81.9 & 55.1 & +26.9 \\
& OpenHands & 79.2 & 54.6 & +24.5 \\
\addlinespace[1pt]
\multirow{2}{*}{Qwen3-Coder-Next} & SWE-Agent & 74.3 & 55.5 & +18.8 \\
& OpenHands & 76.4 & 56.4 & +20.0 \\
\addlinespace[1pt]
\multirow{2}{*}{\makecell[l]{Qwen3-Next-80B-\\A3B-Instruct}} & SWE-Agent & 75.0 & 51.5 & +23.5 \\
& OpenHands & 78.5 & 51.1 & +27.4 \\
\addlinespace[1pt]
\multirow{2}{*}{MiniMax-M2.5} & SWE-Agent & 93.1 & 73.6 & +19.5 \\
& OpenHands & 92.4 & 70.9 & +21.4 \\
\addlinespace[1pt]
\multirow{2}{*}{\makecell[l]{Qwen3-Coder-\\480B-A35B}} & SWE-Agent & 86.1 & 66.1 & +20.0 \\
& OpenHands & 88.2 & 64.3 & +23.9 \\
\midrule
\multicolumn{2}{l}{\textbf{Aggregate}} & \textbf{82.5} & \textbf{59.9} & \textbf{+22.6} \\
\bottomrule
\end{tabular}
}
\end{table}

\FloatBarrier

\subsection{\texorpdfstring{Paired Uncertainty Analysis}{Paired Uncertainty Analysis}}
\label{app:e2e_paired_uncertainty}

\begin{table*}[t]
\centering
\small
\setlength{\tabcolsep}{6pt}
\caption{\textbf{Paired Baseline-to-\textsc{SHERLOC} uncertainty analysis on \textsc{SWE-Bench Verified}}.
Differences and confidence intervals are in percentage points.
Intervals use $10{,}000$ paired bootstrap resamples; $p$-values use continuity-corrected McNemar tests with Holm adjustment across the ten cells.
Bold $p$-values are below $0.05$ after adjustment.}
\label{tab:e2e_paired_uncertainty}
\revtable{%
\begin{tabular}{llrrr}
\toprule
\textbf{Model} & \textbf{Framework} & \textbf{Paired $\Delta$} & \textbf{$95\%$ CI} & \textbf{Holm-adj. $p$} \\
\midrule
\multirow{2}{*}{Qwen3-Coder-30B-A3B}
  & SWE-Agent & $+9.6$ & $[+5.4,+13.8]$ & \textbf{$8.6{\times}10^{-5}$} \\
  & OpenHands & $+3.6$ & $[-0.4,+7.6]$ & $0.277$ \\
\midrule
\multirow{2}{*}{Qwen3-Coder-Next}
  & SWE-Agent & $+8.0$ & $[+3.8,+12.2]$ & \textbf{$0.0022$} \\
  & OpenHands & $+8.6$ & $[+4.2,+13.0]$ & \textbf{$0.0014$} \\
\midrule
\multirow{2}{*}{Qwen3-Next-80B-A3B-Instruct}
  & SWE-Agent & $+10.6$ & $[+6.4,+14.8]$ & \textbf{$2.5{\times}10^{-5}$} \\
  & OpenHands & $+11.8$ & $[+7.8,+16.0]$ & \textbf{$8.2{\times}10^{-7}$} \\
\midrule
\multirow{2}{*}{MiniMax-M2.5}
  & SWE-Agent & $-4.0$ & $[-7.6,-0.4]$ & $0.135$ \\
  & OpenHands & $-5.0$ & $[-8.2,-1.8]$ & \textbf{$0.0168$} \\
\midrule
\multirow{2}{*}{Qwen3-Coder-480B-A35B}
  & SWE-Agent & $0.0$ & $[-3.4,+3.4]$ & $1.000$ \\
  & OpenHands & $+1.0$ & $[-2.6,+4.6]$ & $1.000$ \\
\bottomrule
\end{tabular}
}
\end{table*}

Every condition uses the same $500$ \textsc{SWE-Bench Verified} instances, so we compare Baseline and \textsc{SHERLOC} outcomes pairwise within each backbone--framework cell.
We estimate $95\%$ confidence intervals for the paired resolve-rate difference (\textsc{SHERLOC} minus Baseline) using $10{,}000$ nonparametric bootstrap resamples of benchmark instances~\mbox{\citep{dror2018hitchhikers}}.
We additionally apply continuity-corrected McNemar tests~\mbox{\citep{mcnemar1947note,edwards1948note}} and adjust their $p$-values across the ten cells using Holm's procedure~\mbox{\citep{holm1979simple}}.
The bootstrap intervals are unadjusted.
These analyses quantify uncertainty across benchmark instances, not stochastic run-to-run variation, which would require multi-seed end-to-end evaluations.

\Cref{tab:e2e_paired_uncertainty} shows that \textsc{SHERLOC} improves resolve rate significantly in five of ten cells after correction: Qwen3-Coder-30B-A3B with \textsc{SWE-Agent}, and both frameworks for Qwen3-Coder-Next and Qwen3-Next-80B-A3B-Instruct.
MiniMax-M2.5 is the only backbone with negative point estimates in both frameworks. After Holm correction, the $5.0$~pp OpenHands decrease is significant; the $4.0$~pp \textsc{SWE-Agent} decrease is not, despite its pointwise interval excluding zero.

\FloatBarrier

Relative to unfiltered \textsc{SHERLOC}, QF \textsc{SHERLOC} improves MiniMax-M2.5 by $+6.2$~pp with \textsc{SWE-Agent} ($95\%$ CI $[+3.2,+9.2]$) and $+5.6$~pp with OpenHands ($[+2.6,+8.6]$); both remain significant after Holm correction. Relative to Baseline, its $+2.2$ and $+0.6$~pp differences are not significant.

\section{Localization-Phase Compute Cost for Repair Agents}
\label{app:localization_efficiency}

\Cref{tab:localization_efficiency} reports the per-instance cost of the localization sub-phase of issue resolution on \textsc{SWE-Bench Verified} for \textsc{SWE-Agent} and \textsc{OpenHands} across the $5$ repair models, with and without \textsc{SHERLOC} findings, alongside the cost of the standalone \textsc{SHERLOC} localizer that produces those findings. The localization phase is defined as the prefix of the agent's interaction trace that precedes its first edit action of any kind.

We emphasize that this prefix measures everything the agent does before it starts editing, and that this is \emph{not} the same as ``unconstrained search from scratch'' once findings are injected. Inspection of the traces shows that the standard agent prompt under \textsc{SHERLOC} findings instructs the model to ($i$) orient itself in the repository (working directory, version, layout), ($ii$) read and verify the source file(s) named in the findings, ($iii$) locate and inspect the relevant existing tests, and ($iv$) write a reproduction script before applying any source-level fix. All of these steps appear in the localization phase and are reported by the table; only the actual repair edits and post-edit iterations fall outside it.

\emph{Input tokens} are tokens sent to the model on each call (system prompt plus the full prior conversation, including tool outputs returned in earlier turns); \emph{output tokens} are tokens emitted by the model on each call (assistant text, reasoning, and the function name plus arguments of any emitted tool call). Tool outputs themselves are only counted on the input side of the next call. \emph{Localization total} is the sum of localization input and output; \emph{full-run input/output} are the same metrics summed over the entire interaction (localization plus repair). \emph{Resolved} is the SWE-Bench official resolve rate, included to show that compute reductions from \textsc{SHERLOC} are visible even when resolve rate is comparable: stronger repair models that already solve a high fraction of instances on their own can still benefit on the cost axis.

\paragraph{Per-Backbone Cost-Accuracy Trends.} For Qwen3-Coder-30B-A3B, \textsc{SHERLOC} improves both efficiency and accuracy: paired with \textsc{SWE-Agent}, the effective localization total falls from $101.8$k to $65.8$k tokens while resolve rate increases from $44.7\%$ to $54.0\%$; paired with \textsc{OpenHands}, localization total falls from $361.1$k to $234.1$k tokens while resolve rate increases from $49.4\%$ to $53.0\%$. For MiniMax-M2.5, a much stronger repair model, the picture separates cost from accuracy: \textsc{SHERLOC} lowers localization cost substantially ($116.3$k to $105.6$k with \textsc{SWE-Agent}; $428.3$k to $293.5$k with \textsc{OpenHands}) and lowers full-run token totals, even though resolve rate decreases by about $4$ to $5$ points. Thus, predicted diagnostic context can be useful as a compute-saving device even when it is not accuracy-improving for a strong model. Qwen3-Coder-Next behaves like the $30$B/$80$B mid-tier cases: \textsc{SHERLOC} lowers localization total tokens substantially ($1598$k to $1059$k with \textsc{SWE-Agent}; $1145$k to $727$k with \textsc{OpenHands}) and lifts resolve rate ($45.2\%$ to $53.2\%$ and $44.6\%$ to $53.2\%$), while full-run token totals stay close to baseline ($-$7.3\% / $+$1.4\%). The Qwen3-Coder-480B-A35B row shows a distinct, approximately flat accuracy pattern: \textsc{SHERLOC} lowers localization total tokens substantially in both frameworks ($188.9$k to $64.6$k with \textsc{SWE-Agent}; $472.8$k to $324.6$k with \textsc{OpenHands}) and lowers full-run token totals, while resolve rate is essentially flat ($0.0$\,pp \textsc{SWE-Agent}, $+$1.0\,pp \textsc{OpenHands}). At this model scale, \textsc{SHERLOC} again behaves as a compute-saving device rather than an accuracy-improving one. Qwen3-Next-80B-A3B-Instruct sits between the $30$B and $480$B regimes and behaves like the $30$B case: \textsc{SHERLOC} lowers localization total tokens ($161.4$k to $32.9$k with \textsc{SWE-Agent}; $181.6$k to $105.2$k with \textsc{OpenHands}), lowers full-run token totals ($-$48.4\% with \textsc{SWE-Agent} after adding the localizer cost; $-$33.0\% with \textsc{OpenHands}), and increases resolve rate by $10$--$12$\,pp in both frameworks ($38.8\%$ to $49.4\%$ and $38.6\%$ to $50.4\%$). Like the $30$B model, this is a regime where predicted findings are simultaneously cost-saving and accuracy-improving, even though the localization phase already runs at a relatively low turn count ($9.2$--$9.6$ turns at baseline).

\begin{table*}[htbp]
\centering
\scriptsize
\setlength{\tabcolsep}{3pt}
\renewcommand{\arraystretch}{0.9}
\caption{\textbf{Localization-phase and full-run costs with and without \textsc{SHERLOC} findings}.
Per-instance means over $500$ \textsc{SWE-Bench Verified} tasks; the localization phase is the trace prefix before the first source-edit action.
The ``\textbf{Total pipeline}'' row adds the standalone-localizer cost (top of table) to the agent's own interaction cost with the findings.
$\Delta$ reports percentage change from baseline for cost columns and percentage-point change for resolve rate.}
\label{tab:localization_efficiency}
\begin{tabularx}{\textwidth}{@{}>{\raggedright\arraybackslash}X r r r r r r@{}}
\toprule
\textbf{Setting} & \makecell{\textbf{Loc.}\\\textbf{turns}} & \makecell{\textbf{Loc.}\\\textbf{input}} & \makecell{\textbf{Loc.}\\\textbf{output}} & \makecell{\textbf{Loc.}\\\textbf{total}} & \makecell{\textbf{Full-run}\\\textbf{total}} & \textbf{Resolved} \\
\midrule
\multicolumn{7}{@{}l}{\textbf{Standalone localizer}} \\
\textsc{SHERLOC} (Qwen3-235B) & 4.7 & 9.0k & 19.5k & 28.6k & -- & -- \\
\midrule
\multicolumn{7}{@{}l}{\textbf{Qwen3-Coder-30B-A3B / \textit{SWE-Agent}}} \\
Baseline & 12.3 & 100.5k & 1.3k & 101.8k & 622.4k & 44.7\% \\
\textbf{Total pipeline} & \textbf{10.5} & \textbf{45.4k} & \textbf{20.4k} & \textbf{65.8k} & \textbf{433.6k} & \textbf{54.0\%} \\
$\Delta$ vs baseline & -14.6\% & -54.8\% & +1474.9\% & -35.4\% & -30.3\% & +9.3\,pp \\
\addlinespace[2pt]
\multicolumn{7}{@{}l}{\textbf{Qwen3-Coder-30B-A3B / \textit{OpenHands}}} \\
Baseline & 21.6 & 358.3k & 2.8k & 361.1k & 1630.5k & 49.4\% \\
\textbf{Total pipeline} & \textbf{17.7} & \textbf{212.7k} & \textbf{21.3k} & \textbf{234.1k} & \textbf{1299.4k} & \textbf{53.0\%} \\
$\Delta$ vs baseline & -18.1\% & -40.6\% & +658.7\% & -35.2\% & -20.3\% & +3.6\,pp \\
\midrule
\multicolumn{7}{@{}l}{\textbf{Qwen3-Coder-Next / \textit{SWE-Agent}}} \\
Baseline & 53.3 & 1592.5k & 5.6k & 1598.1k & 3025.9k & 45.2\% \\
\textbf{Total pipeline} & \textbf{45.7} & \textbf{1034.9k} & \textbf{24.0k} & \textbf{1058.9k} & \textbf{2805.6k} & \textbf{53.2\%} \\
$\Delta$ vs baseline & -14.3\% & -35.0\% & +324.5\% & -33.7\% & -7.3\% & +8.0\,pp \\
\addlinespace[2pt]
\multicolumn{7}{@{}l}{\textbf{Qwen3-Coder-Next / \textit{OpenHands}}} \\
Baseline & 40.5 & 1140.8k & 4.6k & 1145.4k & 3221.4k & 44.6\% \\
\textbf{Total pipeline} & \textbf{33.0} & \textbf{703.9k} & \textbf{22.7k} & \textbf{726.6k} & \textbf{3267.8k} & \textbf{53.2\%} \\
$\Delta$ vs baseline & -18.5\% & -38.3\% & +394.5\% & -36.6\% & +1.4\% & +8.6\,pp \\
\midrule
\multicolumn{7}{@{}l}{\textbf{Qwen3-Next-80B-A3B-Instruct / \textit{SWE-Agent}}} \\
Baseline & 9.2 & 160.0k & 1.4k & 161.4k & 617.0k & 38.8\% \\
\textbf{Total pipeline} & \textbf{8.6} & \textbf{41.4k} & \textbf{20.1k} & \textbf{61.5k} & \textbf{318.3k} & \textbf{49.4\%} \\
$\Delta$ vs baseline & -6.5\% & -74.1\% & +1373.0\% & -61.9\% & -48.4\% & +10.6\,pp \\
\addlinespace[2pt]
\multicolumn{7}{@{}l}{\textbf{Qwen3-Next-80B-A3B-Instruct / \textit{OpenHands}}} \\
Baseline & 9.6 & 179.6k & 2.0k & 181.6k & 626.8k & 38.6\% \\
\textbf{Total pipeline} & \textbf{10.9} & \textbf{112.4k} & \textbf{21.4k} & \textbf{133.8k} & \textbf{420.1k} & \textbf{50.4\%} \\
$\Delta$ vs baseline & +13.5\% & -37.4\% & +965.4\% & -26.3\% & -33.0\% & +11.8\,pp \\
\midrule
\multicolumn{7}{@{}l}{\textbf{MiniMax-M2.5 / \textit{SWE-Agent}}} \\
Baseline & 11.3 & 114.3k & 2.0k & 116.3k & 1556.3k & 74.4\% \\
\textbf{Total pipeline} & \textbf{12.5} & \textbf{84.8k} & \textbf{20.8k} & \textbf{105.6k} & \textbf{1206.1k} & \textbf{70.4\%} \\
$\Delta$ vs baseline & +10.6\% & -25.8\% & +918.5\% & -9.2\% & -22.5\% & -4.0\,pp \\
\addlinespace[2pt]
\multicolumn{7}{@{}l}{\textbf{MiniMax-M2.5 / \textit{OpenHands}}} \\
Baseline & 22.1 & 423.7k & 4.6k & 428.3k & 1591.3k & 72.2\% \\
\textbf{Total pipeline} & \textbf{19.1} & \textbf{270.8k} & \textbf{22.6k} & \textbf{293.5k} & \textbf{1395.0k} & \textbf{67.2\%} \\
$\Delta$ vs baseline & -13.6\% & -36.1\% & +394.3\% & -31.5\% & -12.3\% & -5.0\,pp \\
\midrule
\multicolumn{7}{@{}l}{\textbf{Qwen3-Coder-480B-A35B / \textit{SWE-Agent}}} \\
Baseline & 15.2 & 187.6k & 1.3k & 188.9k & 1157.0k & 63.0\% \\
\textbf{Total pipeline} & \textbf{10.8} & \textbf{44.5k} & \textbf{20.0k} & \textbf{64.6k} & \textbf{596.5k} & \textbf{63.0\%} \\
$\Delta$ vs baseline & -28.9\% & -76.3\% & +1408.8\% & -65.8\% & -48.4\% & +0.0\,pp \\
\addlinespace[2pt]
\multicolumn{7}{@{}l}{\textbf{Qwen3-Coder-480B-A35B / \textit{OpenHands}}} \\
Baseline & 24.5 & 470.0k & 2.8k & 472.8k & 2067.5k & 61.6\% \\
\textbf{Total pipeline} & \textbf{21.4} & \textbf{303.0k} & \textbf{21.6k} & \textbf{324.6k} & \textbf{1862.9k} & \textbf{62.6\%} \\
$\Delta$ vs baseline & -12.7\% & -35.5\% & +669.5\% & -31.3\% & -9.9\% & +1.0\,pp \\
\bottomrule
\end{tabularx}
\end{table*}

\begin{table*}[htbp]
\centering
\scriptsize
\setlength{\tabcolsep}{3pt}
\renewcommand{\arraystretch}{1.02}
\caption{\textbf{Localization-phase efficiency with Qwen3-30B \textsc{SHERLOC} findings}.
Per-instance means over $500$ \textsc{SWE-Bench Verified} tasks for repair runs that receive findings produced by Qwen3-30B-A3B-Thinking-2507.
``Qwen3-30B total'' columns add the standalone Qwen3-30B \textsc{SHERLOC} localizer cost ($7.2$ turns; $10.9$k input + $35.1$k output = $46.0$k total tokens) to the repair-agent run.
Resolve rates are shown as baseline $\rightarrow$ with Qwen3-30B \textsc{SHERLOC} findings.}
\label{tab:localization_efficiency_qwen30_findings}
\begin{tabularx}{\textwidth}{@{}l l r r r r r r r@{}}
\toprule
\textbf{Model} & \textbf{Framework} & \makecell{\textbf{Baseline}\\\textbf{loc. tok}} & \makecell{\textbf{Qwen3-30B}\\\textbf{loc. tok}} & \makecell{\boldmath$\Delta$\\\textbf{loc.}} & \makecell{\textbf{Baseline}\\\textbf{full tok}} & \makecell{\textbf{Qwen3-30B}\\\textbf{full tok}} & \makecell{\boldmath$\Delta$\\\textbf{full}} & \textbf{Resolved} \\
\midrule
\multirow{2}{*}{Qwen3-Coder-30B-A3B}
  & SWE-Agent & 101.8k & 85.4k & $-$16.1\% & 622.4k & 574.4k & $-$7.7\% & 44.7\% $\rightarrow$ 43.8\% \\
  & OpenHands & 361.1k & 246.5k & $-$31.7\% & 1630.5k & 1351.8k & $-$17.1\% & 49.4\% $\rightarrow$ 42.2\% \\
\midrule
\multirow{2}{*}{Qwen3-Coder-Next}
  & SWE-Agent & 1598.1k & 1224.6k & $-$23.4\% & 3025.9k & 2888.2k & $-$4.6\% & 45.2\% $\rightarrow$ 49.2\% \\
  & OpenHands & 1145.4k & 840.5k & $-$26.6\% & 3221.4k & 3197.9k & $-$0.7\% & 44.6\% $\rightarrow$ 45.4\% \\
\bottomrule
\end{tabularx}
\end{table*}

\paragraph{Patch Production: No-Patch and Apply-Failure Rates.}
The no-patch and apply-failure rates from the completed runs reinforce the efficiency interpretation above.
For Qwen3-Coder-Next/OpenHands, the baseline has $26.6\%$ no-patch and $27.0\%$ apply-failure rates; with \textsc{SHERLOC} findings these fall to $16.2\%$ and $16.6\%$, respectively, paralleling the resolve-rate lift.
For strong models, no-patch and apply-failure rates are already low, so the main measured effect remains search and token cost rather than patch production.
\Cref{tab:localization_efficiency_qwen30_findings} shows the same pattern with the smaller Qwen3-30B localizer: localization-phase tokens decrease in every measured cell.

\section{Token Cost Under the Masked-Description Control}
\label{app:masked_masking_tokens}

\Cref{tab:masked_masking_tokens} checks whether masking repository/file hints in the problem statement makes repair agents spend more tokens on localization.
Across completed baseline--masked pairs, the effect is mostly small and mixed: with the exception of the Qwen3-Coder-480B-A35B/SWE-Agent outlier ($+23.0\%$), localization-token deltas range from $-12.0\%$ to $+10.9\%$.
Thus, masking problem-statement identifiers generally does not force substantially more localization effort; agents largely follow similar exploration routines, or infer structure from the remaining context.

\begin{table*}[htbp]
\centering
\scriptsize
\setlength{\tabcolsep}{3pt}
\caption{\textbf{Baseline vs.\ masked-description token cost}.
``Loc.'' is the localization phase (the prefix of the trajectory before the first source-edit action).
Per-instance means; deltas are masked relative to baseline.}
\label{tab:masked_masking_tokens}
\resizebox{\textwidth}{!}{%
\begin{tabular}{llrrrrrr}
\toprule
\textbf{Model} & \textbf{Framework} & \textbf{Loc. turns} & \textbf{Loc. tokens} & \textbf{$\Delta$ loc. tok} & \textbf{Full-run tokens} & \textbf{$\Delta$ full tok} & \textbf{Resolved} \\
\midrule
\multirow{2}{*}{Qwen3-Coder-30B-A3B}
  & SWE-Agent & 12.3 $\rightarrow$ 12.8 & 101.8k $\rightarrow$ 111.4k & +9.4\% & 622.4k $\rightarrow$ 997.4k & +60.2\% & 44.7\% $\rightarrow$ 45.4\% \\
  & OpenHands & 21.6 $\rightarrow$ 21.7 & 361.1k $\rightarrow$ 368.2k & +2.0\% & 1630.5k $\rightarrow$ 1644.9k & +0.9\% & 49.4\% $\rightarrow$ 46.0\% \\
\midrule
\multirow{2}{*}{Qwen3-Coder-Next}
  & SWE-Agent & 53.3 $\rightarrow$ 52.7 & 1598.1k $\rightarrow$ 1599.8k & +0.1\% & 3025.9k $\rightarrow$ 2871.4k & -5.1\% & 45.2\% $\rightarrow$ 41.2\% \\
  & OpenHands & 40.5 $\rightarrow$ 40.4 & 1145.4k $\rightarrow$ 1131.1k & -1.2\% & 3221.4k $\rightarrow$ 3125.6k & -3.0\% & 44.6\% $\rightarrow$ 44.6\% \\
\midrule
\multirow{2}{*}{Qwen3-Next-80B-A3B-Instruct}
  & SWE-Agent & 9.2 $\rightarrow$ 9.0 & 161.4k $\rightarrow$ 142.1k & -12.0\% & 617.0k $\rightarrow$ 607.1k & -1.6\% & 38.8\% $\rightarrow$ 41.6\% \\
  & OpenHands & 9.6 $\rightarrow$ 9.9 & 181.6k $\rightarrow$ 190.9k & +5.1\% & 626.8k $\rightarrow$ 697.9k & +11.3\% & 38.6\% $\rightarrow$ 38.2\% \\
\midrule
\multirow{2}{*}{MiniMax-M2.5}
  & SWE-Agent & 11.3 $\rightarrow$ 12.3 & 116.3k $\rightarrow$ 129.0k & +10.9\% & 1556.3k $\rightarrow$ 1642.7k & +5.6\% & 74.4\% $\rightarrow$ 74.5\% \\
  & OpenHands & 22.1 $\rightarrow$ 22.5 & 428.3k $\rightarrow$ 437.7k & +2.2\% & 1591.3k $\rightarrow$ 1586.6k & -0.3\% & 72.2\% $\rightarrow$ 71.8\% \\
\midrule
\multirow{2}{*}{Qwen3-Coder-480B-A35B}
  & SWE-Agent & 15.2 $\rightarrow$ 16.1 & 188.9k $\rightarrow$ 232.4k & +23.0\% & 1157.0k $\rightarrow$ 1269.0k & +9.7\% & 63.0\% $\rightarrow$ 61.8\% \\
  & OpenHands & 24.5 $\rightarrow$ 24.6 & 472.8k $\rightarrow$ 470.5k & -0.5\% & 2067.5k $\rightarrow$ 2066.1k & -0.1\% & 61.6\% $\rightarrow$ 62.8\% \\
\bottomrule
\end{tabular}%
}
\end{table*}

\section{File-Level Localization: Baseline vs.\ \textsc{SHERLOC}}
\label{app:baseline_vs_sherloc_localization}

For each \textsc{SWE-Bench Verified} instance we extract the implicit localization output of baseline and masked agent runs, i.e., the set of \emph{existing source files} that the agent actually modifies in its model patch, excluding agent-created reproduction, debug, or test scripts (any hunk introducing a new file is filtered out by checking for \texttt{new file mode} or \texttt{--- /dev/null}).
We compare these file sets to \textsc{SHERLOC}'s predicted files and to the gold patch files.
\Cref{tab:baseline_vs_sherloc_localization} reports the full per-instance file-level localization breakdown that complements the compute and resolve-rate view in \Cref{tab:localization_efficiency}: Hit@$1$, macro precision/recall, and macro F$1$ over all $500$ tasks.
Instances where the agent never touches an existing source file contribute zero, so the metric jointly rewards correct file selection and the willingness to commit to one.

\begin{table*}[htbp]
\centering
\scriptsize
\setlength{\tabcolsep}{4pt}
\caption{\textbf{File-level localization quality on \textsc{SWE-Bench Verified}}.
\textsc{SHERLOC}'s predicted files vs.\ the existing-source files modified by baseline and masked agent runs; higher is better.
\textbf{Evaluable\,(n)} is the number of instances where the method emits at least one file (\textsc{SHERLOC} predicts files for almost all instances; baseline/masked counts equal the rate at which the agent commits a real source edit).
Hit@$1$, Recall, Precision, and F$1$ are per-instance means over all $500$ instances, with non-evaluable instances contributing zero.}
\label{tab:baseline_vs_sherloc_localization}
\resizebox{\textwidth}{!}{%
\begin{tabular}{ll l rrrr r}
\toprule
\textbf{Model} & \textbf{Framework} & \textbf{Method} & \textbf{Hit@$1$} & \textbf{Recall} & \textbf{Precision} & \textbf{F$1$} & \textbf{Evaluable\,(n)} \\
\midrule
Qwen3-235B-A22B-Thinking-2507 & \textsc{SHERLOC} &  & \textbf{0.884} & \textbf{0.813} & \textbf{0.876} & \textbf{0.835} & 498 / 500 \\
\midrule
\multirow{4}{*}{Qwen3-Coder-30B-A3B}
  & \multirow{2}{*}{SWE-Agent}
                  & Baseline & 0.772 & 0.730 & 0.763 & 0.734 & 437 / 500 \\
  &              & Masked & 0.788 & 0.735 & 0.781 & 0.743 & 457 / 500 \\
  & \multirow{2}{*}{OpenHands}
                  & Baseline & 0.752 & 0.759 & 0.748 & 0.730 & 461 / 500 \\
  &              & Masked & 0.734 & 0.738 & 0.725 & 0.708 & 456 / 500 \\
\midrule
\multirow{4}{*}{Qwen3-Coder-Next}
  & \multirow{2}{*}{SWE-Agent}
                  & Baseline & 0.562 & 0.553 & 0.551 & 0.542 & 305 / 500 \\
  &              & Masked & 0.494 & 0.483 & 0.485 & 0.474 & 271 / 500 \\
  & \multirow{2}{*}{OpenHands}
                  & Baseline & 0.562 & 0.584 & 0.543 & 0.547 & 357 / 500 \\
  &              & Masked & 0.562 & 0.573 & 0.540 & 0.538 & 350 / 500 \\
\midrule
\multirow{4}{*}{Qwen3-Next-80B-A3B-Instruct}
  & \multirow{2}{*}{SWE-Agent}
                  & Baseline & 0.760 & 0.721 & 0.761 & 0.728 & 441 / 500 \\
  &              & Masked & 0.766 & 0.726 & 0.766 & 0.731 & 447 / 500 \\
  & \multirow{2}{*}{OpenHands}
                  & Baseline & 0.752 & 0.776 & 0.752 & 0.738 & 480 / 500 \\
  &              & Masked & 0.744 & 0.775 & 0.734 & 0.731 & 479 / 500 \\
\midrule
\multirow{4}{*}{MiniMax-M2.5}
  & \multirow{2}{*}{SWE-Agent}
                  & Baseline & \textbf{0.896} & \textbf{0.864} & \textbf{0.888} & \textbf{0.863} & 489 / 500 \\
  &              & Masked & 0.886 & 0.848 & 0.880 & 0.851 & 461 / 500 \\
  & \multirow{2}{*}{OpenHands}
                  & Baseline & 0.824 & 0.846 & 0.818 & 0.809 & 493 / 500 \\
  &              & Masked & 0.834 & 0.855 & 0.823 & 0.816 & 487 / 500 \\
\midrule
\multirow{4}{*}{Qwen3-Coder-480B-A35B}
  & \multirow{2}{*}{SWE-Agent}
                  & Baseline & 0.868 & 0.834 & 0.844 & 0.821 & 491 / 500 \\
  &              & Masked & 0.862 & 0.833 & 0.839 & 0.818 & 488 / 500 \\
  & \multirow{2}{*}{OpenHands}
                  & Baseline & 0.816 & 0.837 & 0.773 & 0.776 & 490 / 500 \\
  &              & Masked & 0.812 & 0.837 & 0.785 & 0.785 & 490 / 500 \\
\bottomrule
\end{tabular}%
}
\end{table*}

\begin{table*}[htbp]
\centering
\scriptsize
\setlength{\tabcolsep}{4pt}
\caption{\textbf{Standalone Qwen3-30B \textsc{SHERLOC} localization vs.\ baseline repair-agent traces}.
The injected-finding repair runs in \Cref{tab:localization_efficiency_qwen30_findings} do not receive a separate localization score; their findings are supplied by the standalone \textsc{SHERLOC} row.}
\label{tab:qwen30_findings_localization}
\resizebox{\textwidth}{!}{%
\begin{tabular}{ll l rrrr r}
\toprule
\textbf{Model} & \textbf{Framework} & \textbf{Method} & \textbf{Hit@$1$} & \textbf{Recall} & \textbf{Precision} & \textbf{F$1$} & \textbf{Evaluable\,(n)} \\
\midrule
Qwen3-30B-A3B-Thinking-2507 & \textsc{SHERLOC} &  & \textbf{0.786} & \textbf{0.751} & \textbf{0.785} & \textbf{0.750} & 497 / 500 \\
\midrule
\multirow{2}{*}{Qwen3-Coder-30B-A3B}
  & SWE-Agent   & Baseline & 0.772 & 0.730 & 0.763 & 0.734 & 437 / 500 \\
  & OpenHands   & Baseline & 0.752 & 0.759 & 0.748 & 0.730 & 461 / 500 \\
\midrule
\multirow{2}{*}{Qwen3-Coder-Next}
  & SWE-Agent   & Baseline & 0.562 & 0.553 & 0.551 & 0.542 & 305 / 500 \\
  & OpenHands   & Baseline & 0.562 & 0.584 & 0.543 & 0.547 & 357 / 500 \\
\bottomrule
\end{tabular}%
}
\end{table*}

\paragraph{Reading.}
\Cref{tab:localization_efficiency} reports compute and resolve rate; this table provides the matched file-level localization-metric breakdown.
\textsc{SHERLOC} dominates baseline/masked agent localization for the weaker models, while MiniMax-M2.5 already localizes near saturation.
Masked ($\approx$ baseline) values across rows indicate that masking file/repo identifiers in the problem statement is not a strong contamination control here: capable agents recover localization quality without those hints.
\Cref{tab:qwen30_findings_localization} isolates the Qwen3-30B localizer used for the smaller-finding runs: it is less accurate than Qwen3-235B \textsc{SHERLOC}, but remains stronger than the baseline localization traces for the weaker repair-agent settings.

\end{document}